\RequirePackage{fix-cm}
\RequirePackage[T1]{fontenc}
\documentclass[letterpaper,10pt,conference]{ieeeconf}
\newif\ifextended\extendedtrue
\IEEEoverridecommandlockouts
\usepackage[T1]{fontenc}
\usepackage{amsmath,amssymb,graphicx,booktabs,array,tabularx,cite}
\usepackage[table]{xcolor}
\usepackage{flafter,placeins,ragged2e}
\usepackage{siunitx}
\definecolor{rowgray}{HTML}{F3F3F3}
\usepackage[hidelinks]{hyperref}
\hypersetup{pdftitle={Measuring the Value of One Shared Image at Robot Handoffs in Vision-and-Language Navigation},pdfauthor={Yuze Lu},pdfsubject={Paired measurements of robot team success and communication cost}}
\newcommand{\pp}{\,\mathrm{pp}}
\newcommand{\arm}[1]{\textsf{#1}}
\newcommand{\fig}[1]{Fig.~\ref{#1}}
\newcommand{\tab}[1]{Table~\ref{#1}}
\newcommand{\protocolfig}{%
\begin{figure}[t]
\centering\includegraphics[width=\columnwidth]{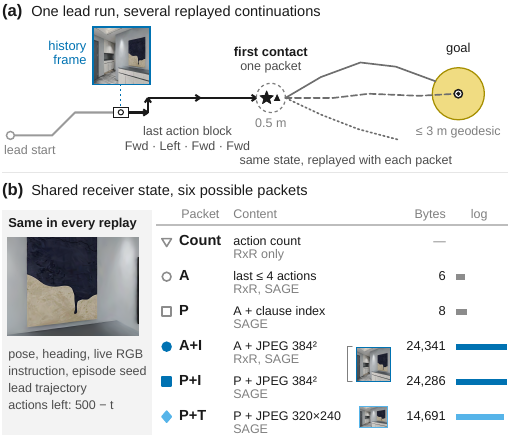}
\caption{A replayable handoff. (a) The camera marks the lead robot's view before its last action block. At first contact within 0.5\,m geodesic, each message produces a continuation from the same state (schematic). (b) Pose, heading, current view, instruction, seed, lead trajectory and remaining actions are shared. Packet contents and application bytes are from SAGE 841498/53. Bars use a log scale from 1~B to 30~kB. \arm{A+I} and \arm{P+I} carry the same JPEG. \arm{P+T} carries a thumbnail of the same camera frame.}
\label{f:handoff}\end{figure}}

\newcommand{\effectsfig}{%
\begin{figure}[t]
\centering\includegraphics[width=\columnwidth]{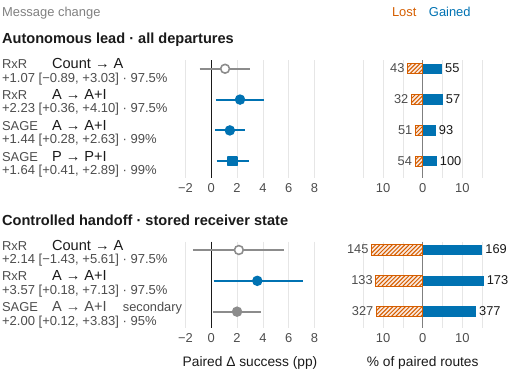}
\caption{Paired effects at the same handoff, right message minus left. Autonomous rows include all departures. Controlled rows use the stored receiver state. SAGE uses whole-scene two-sided 99\% intervals for its prespecified contrasts and 95\% for the secondary comparison. RxR uses within-building 97.5\% intervals. Both Count-to-\arm{A} intervals cross zero. Lost and gained episodes share one percentage scale across panels, with pooled counts at the bar ends. The difference between the gained and lost percentages gives the route-pooled effect. The SAGE forest estimates are scene-equal.}
\label{f:effects}\end{figure}}

\newcommand{\geometryfig}{%
\begin{figure}[t]
\centering\includegraphics[width=\columnwidth]{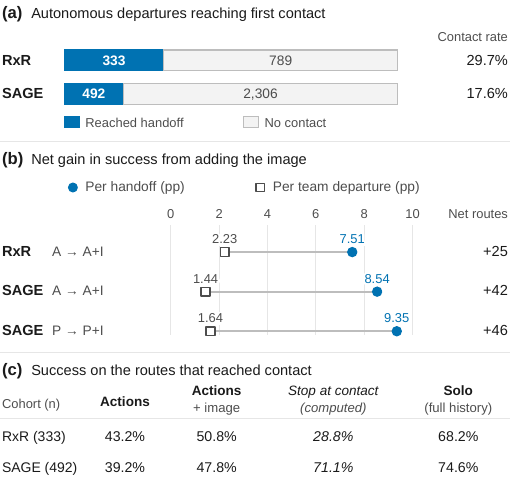}
\caption{How often a handoff occurs and how its gain carries through to the team. (a) The bars cover all 1,122 RxR and 2,798 SAGE departures. (b) Circles show route-pooled differences among realized handoffs. Squares use all departures, counting missed contacts as zero in both arms. Team estimates pool routes in RxR and weight the 95 scenes equally in SAGE. The last column counts rescues minus reversals. (c) All four rates use the same realized contacts within each cohort. \emph{Stop at contact (computed)} scores the receiver pose against the 3\,m success radius. Solo is the empirical full-history ceiling.}
\label{f:geometry}\end{figure}}

\newcommand{\casesfig}{%
\begin{figure*}[t]
\centering\includegraphics[width=\textwidth]{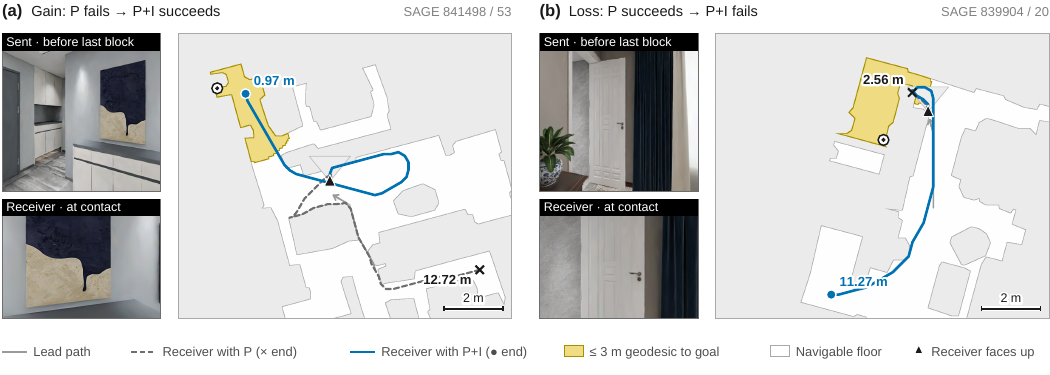}
\caption{One gain and one loss for \arm{P} versus \arm{P+I}. Each panel pairs the transmitted frame and the shared receiver view with both logged paths, projected horizontally. The receiver initially faces up and the two maps share one scale. Shading marks reachable floor within 3\,m geodesic of the goal. In (b), the receiver starts 1.78\,m from the goal in a straight line and 3.36\,m along the navigation mesh. A wall separates the starting pose from the shaded region. For each outcome change, we selected the median \arm{P+I} packet size among departures with finite nominal handoff distance above 3\,m, breaking ties by episode ID.}
\label{f:cases}\end{figure*}}

\newcommand{\tfont}{\fontsize{7.5}{9}\selectfont}
\newcommand{\tnote}[1]{\par\smallskip\parbox{\columnwidth}{\raggedright\fontsize{7}{8.4}\selectfont #1}}
\newcommand{\field}[1]{\shortstack[l]{#1}}
\newcommand{\cohortstable}{%
\begin{table*}[t]
\caption{Cohorts and primary comparisons. Effect directions are given as comparison minus reference.}\label{t:cohorts}
\centering\small\setlength{\tabcolsep}{4pt}
\begin{tabularx}{\textwidth}{@{}llrr>{\raggedright\arraybackslash}X>{\raggedright\arraybackslash}X@{}}
\toprule
\multicolumn{6}{@{}l}{\textbf{(a) Cohorts and contact opportunity}}\\
Cohort & Environments & Routes & Contacts (\%) & Messages & Analysis unit and CI \\
\midrule
\multicolumn{6}{@{}l}{\emph{Protocol development, controlled placement}}\\
R2R & 11 buildings & 150 & --- & None, Count, A, A+I & Paired placements \\
\addlinespace[2pt]
\multicolumn{6}{@{}l}{\emph{Formal evaluation, autonomous lead}}\\
RxR-EN & Same 11 buildings & 1,122 & 333 (29.7) & Count, A, A+I & Route-pooled, 97.5\% \\
SAGE & 95 scenes & 2,798 & 492 (17.6) & A, A+I, P, P+I, P+T, selector & Scene-equal, 99\% \\
\midrule
\multicolumn{2}{@{}l}{Total autonomous departures} & \textbf{3,920} & 825 & & \\
\bottomrule\end{tabularx}
\smallskip
\begin{tabular*}{0.99\textwidth}{@{\extracolsep{\fill}}llrlrl@{}}
\toprule
\multicolumn{6}{@{}l}{\textbf{(b) Primary success differences in the two prespecified families}}\\
Family & Comparison $-$ reference & $n$ & Population and weighting & CI & $\Delta$ success in pp [CI] \\
\midrule
Original SAGE & A+I $-$ A & 2,798 & Departures, scene-equal & 99\% & $+1.44\ [ +0.28,+2.63]$\\
Original SAGE & P+I $-$ P & 2,798 & Departures, scene-equal & 99\% & $+1.64\ [ +0.41,+2.89]$\\
Original SAGE & Selector $-$ global random & 2,798 & Departures, scene-equal & 99\% & $+0.32\ [ -0.11,+0.78]$\\
Original SAGE & Selector $-$ fixed thumbnail (P+T) & 2,798 & Departures, scene-equal & 99\% & $+0.22\ [ -0.72,+1.19]$\\
Original SAGE & P $-$ (A+I) & 2,798 & Departures, scene-equal & 99\% & $-1.25\ [ -2.53,-0.04]$\\
\midrule
Revision N1 & Matched $-$ donor image & 492 & SAGE handoffs, route-pooled & 98.75\% & $+5.28\ [ -1.96,+12.36]$\\
Revision N2 & $(P_{\rm major}+{\rm I})-P_{\rm major}$ & 492 & SAGE handoffs, route-pooled & 98.75\% & $+8.94\ [ +2.80,+14.89]$\\
Revision N3 & $(P_{\rm ref}+{\rm I})-P_{\rm ref}$ & 1,122 & RxR placements, route-pooled & 98.75\% & $+2.32\ [ -1.60,+6.15]$\\
Revision N4 & P+I $-$ P, starts outside 3\,m & 1,136 & SAGE placements, route-pooled & 98.75\% & $+1.50\ [ -2.20,+5.17]$\\
\bottomrule
\end{tabular*}
\par\smallskip\parbox{0.98\textwidth}{\footnotesize Two additional SAGE scenes served for engineering. Of 100 formal candidates, 95 supplied 12--30 eligible departures each. Panel (b) lists the original five SAGE and four revision primary comparisons in their prespecified directions. Missed contacts score zero on team success. A sends actions, P adds estimated progress, $P_{\rm major}$ adds a constant first-clause index and $P_{\rm ref}$ retains unknowns when a unique reference index is unavailable. I is the pre-block image. The original RxR team comparisons, a separate two-contrast family, appear in Fig.~\ref{f:effects}. Formal evaluation comprised 12,421 original receiver executions and 16,408 registered revision executions. The latter include 6,330 baseline reruns. Controlled R2R development placements and the supplementary Solo comparison are separate.}
\end{table*}}

\newcommand{\idash}{\multicolumn{1}{r}{\textit{---}}}
\newcommand{\cbhead}[1]{\multicolumn{\ifextended6\else5\fi}{@{}l}{\emph{#1}}}
\newcommand{\costtable}{%
\begin{table}[t]
\caption{Communication cost and team success in SAGE (95 scenes, 2,798 departures, 492 contacts)}\label{t:cost}
\centering\tfont\setlength{\tabcolsep}{1.7pt}
\ifextended
\begin{tabularx}{\columnwidth}{@{}>{\raggedright\arraybackslash}X S[table-format=3.2] S[table-format=3.2] S[table-format=2.2] S[table-format=4.2] r@{}}
\toprule
Policy & {Images} & {\field{Team\\successes}} & {SR (\%)} & {B/departure} & {\field{Total\\bytes}}\\
\midrule
\cbhead{Measured continuations}\\
\arm{P} (short packet) &0 &198 &7.13 &1.41 &\num{3936}\\
\arm{P+I} (model-view image) &492 &244 &8.77 &4715.77 &\num{13194720}\\
\arm{P+T} (thumbnail) &492 &240 &8.52 &2827.51 &\num{7911375}\\
\rowcolor{rowgray}Selector (frozen) &423 &244 &8.75 &3735.00 &\num{10450537}\\
\addlinespace[2pt]\cbhead{Expected at the selector's byte expenditure}\\
\itshape Random, global &\itshape 389.65 &\itshape 234.43 &\itshape 8.43 &\itshape 3735.00 &\textit{\num{10450537}}\\
\rowcolor{rowgray}\itshape Random, size-binned &\itshape 420.70 &\itshape 244.31 &\itshape 8.76 &\itshape 3735.00 &\textit{\num{10450537}}\\
\addlinespace[2pt]\cbhead{Computed reference}\\
\itshape Stop at contact &\idash &\itshape 350 &\itshape 12.50 &\idash &\textit{---}\\
\bottomrule\end{tabularx}
\else
\begin{tabularx}{\columnwidth}{@{}>{\raggedright\arraybackslash}X S[table-format=3.2] S[table-format=3.2] S[table-format=2.2] S[table-format=4.2]@{}}
\toprule
Policy & {Images} & {\field{Team\\successes}} & {SR (\%)} & {B/departure}\\
\midrule
\cbhead{Measured continuations}\\
\arm{P} (short packet) &0 &198 &7.13 &1.41\\
\arm{P+I} (model-view image) &492 &244 &8.77 &4716\\
\arm{P+T} (thumbnail) &492 &240 &8.52 &2828\\
\rowcolor{rowgray}Selector (frozen) &423 &244 &8.75 &3735\\
\addlinespace[2pt]\cbhead{Expected at the selector's byte expenditure}\\
\itshape Random, global &\itshape 389.65 &\itshape 234.43 &\itshape 8.43 &\itshape 3735\\
\rowcolor{rowgray}\itshape Random, size-binned &\itshape 420.70 &\itshape 244.31 &\itshape 8.76 &\itshape 3735\\
\addlinespace[2pt]\cbhead{Computed reference}\\
\itshape Stop at contact &\idash &\itshape 350 &\itshape 12.50 &\idash\\
\bottomrule\end{tabularx}
\fi
\tnote{Team successes count completed episodes. SR weights scenes equally across all departures, so identical pooled success counts can give different scene-equal rates. A missed contact costs 0~B. The selector uses its chosen measured continuation. Random rows weight both continuations by send probability. The reference scores an unconditional stop from the recorded pose. Dashes indicate unassigned costs.}
\end{table}}

\newcommand{\SOneSoloTable}{%
\begin{table*}[t]
\caption{Receiver success at realized handoffs and over all departures. Solo is the empirical full-history ceiling. Measured S1 contrasts have pointwise 95\% intervals.}
\label{t:solo_s1}\centering\fontsize{8}{9.5}\selectfont\setlength{\tabcolsep}{3pt}
\begin{tabular*}{\textwidth}{@{\extracolsep{\fill}}llrrcrrr@{}}\toprule
Cohort & Condition & $n$ & SR (\%) & $\Delta$ vs Solo (pp) [95\% CI] & \shortstack{Mean NE\\(m)} & \shortstack{Mean suffix\\SPL} & \shortstack{Median suffix\\actions} \\ \midrule
SAGE & A & 492 & 39.23 & $-35.37\;[-41.26,-29.22]$ & 4.26 & 0.244 & 38 \\
SAGE & A+I & 492 & 47.76 & $-26.83\;[-32.08,-21.70]$ & 3.80 & 0.311 & 32 \\
SAGE & \textit{Stop at contact (computed)} & 492 & \textit{71.14} & --- & --- & --- & --- \\
SAGE & Solo & 492 & 74.59 & ref. & 2.31 & 0.638 & 12 \\
SAGE & P & 492 & 40.24 & $-34.35\;[-40.41,-28.08]$ & 4.20 & 0.252 & 38 \\
SAGE & P+I & 492 & 49.59 & $-25.00\;[-30.44,-19.60]$ & 3.70 & 0.325 & 32 \\
\addlinespace
RxR & A & 333 & 43.24 & $-24.92\;[-30.75,-18.91]$ & 7.03 & 0.307 & 76 \\
RxR & A+I & 333 & 50.75 & $-17.42\;[-23.37,-11.38]$ & 5.86 & 0.414 & 62 \\
RxR & \textit{Stop at contact (computed)} & 333 & \textit{28.83} & --- & --- & --- & --- \\
RxR & Solo & 333 & 68.17 & ref. & 3.49 & 0.655 & 35 \\
\bottomrule\end{tabular*}\vspace{4pt}
\begin{tabular*}{\textwidth}{@{\extracolsep{\fill}}llrrrrr@{}}\toprule
\multicolumn{7}{@{}l}{\textbf{Success over all departures (\%).} SAGE weights scenes equally and RxR pools routes.} \\
Cohort & Departures & Solo & A & A+I & P & P+I \\ \midrule
SAGE & 2,798 & 31.57 & 6.94 & 8.38 & 7.13 & 8.77 \\
RxR & 1,122 & 51.78 & 12.83 & 15.06 & --- & --- \\
\bottomrule\end{tabular*}
\par\smallskip\parbox{0.98\textwidth}{\footnotesize Suffix metrics start at the actual contact pose of the lead for Solo and the actual handoff pose of the receiver for each message. NE and suffix SPL are means, whereas suffix actions are medians. All SAGE contact metrics use 492 routes. RxR SR and action summaries use 333 contacts, while NE and SPL means use 318 routes with finite distances. Among contacts, full-run Solo mean SPL and median actions are 0.612 and 54 in SAGE, and 0.623 and 86 in RxR. The italic stop row scores only the receiver pose at contact. Its SR includes all contacts, including 15 unreachable RxR starts. Differences for measured packets are relative to Solo on the same contacts. In the lower panel, missed contacts score zero for the required handoff and the lead keeps its observed outcome for the full-history reference. P was evaluated in SAGE.}
\end{table*}%
}

\newcommand{\SOneNewSoloRuns}{825}

\newcommand{\SOneSoloGPUHours}{5.803}

\title{\LARGE\bf Measuring the Value of One Shared Image\\ at Robot Handoffs in Vision-and-Language Navigation}
\author{\authorblockN{Yuze Lu}\authorblockA{Tsinghua University}}
\begin{document}
\bstctlcite{bstcontrol}
\maketitle
\thispagestyle{plain}\pagestyle{plain}\suppressfloats[t]
\begin{abstract}
\ifextended
When the lead robot cannot continue, what should it send the teammate that must finish the route? We replay a frozen receiver at each contact under alternative messages. Across 3,920 autonomous departures in 106 environments, adding the view before the lead's last action block to its executed actions raised success by 7.51 percentage points (pp) per RxR handoff and 8.54~pp per SAGE handoff. Per departure the gains were 2.23 and 1.44~pp, respectively. The lead continuing in simulation defines the empirical full-history ceiling (Solo). At contact, the lead had accumulated a median of 41 frames. Sending one frame when available recovered 24\% of the full-history advantage over actions alone in SAGE and 30\% in RxR. Over an estimated progress cue in SAGE, the image rescued 100 routes and reversed 54 successes. Near-goal receivers finished outside less often (63.1\% to 51.7\%, post hoc). In secondary comparisons the earlier view outperformed the contact view by 6--7~pp per handoff. Exact-route specificity and image gains at controlled starts with mostly unknown progress or outside the goal region remained unresolved. Gains were detectable from about 7\,kB per handoff. Random transmission matched within size bins nearly equaled the learned sender's team success. At a required handoff, a view of the approach to the last maneuver helps the receiver finish.
\fi
\end{abstract}

\section{Introduction}
A delivery can pass from a truck to a drone or between robots assigned to different legs~\cite{mathew,parcel}. We study the case in which the lead cannot finish a navigation instruction. Its battery may be depleted, the next segment may require another platform's capabilities, or the team may have assigned that segment to a partner. At contact the receiver must take over. The communication decision is what to send it from the stretch of route it has never seen.

Consider a lead robot that turns left and meets its teammate in a corridor. The receiver can see the corridor ahead and can receive the exact turn sequence in a few bytes. A photograph from just before the turn shows the approach, but costs thousands of times more to transmit. Does that view change which routes the team completes?

Navigation policies already use their own visual history~\cite{stream}, and cooperating navigators can share peer observations~\cite{co}. We isolate the value of a message by running the lead once and replaying a frozen receiver under each candidate packet at the same physical contact (\fig{f:handoff}). All packet branches share the receiver's view, residual budget and instruction. For an empirical full-history ceiling (Solo), we allow the original lead to finish in simulation with its own history. In the deployment task, only the receiver can finish the route.

\protocolfig

\ifextended
The lead had accumulated a median of 41 camera frames by first contact. The median was 40 in SAGE (interquartile range [IQR], 23--57) and 43 in RxR (26--62). Sending one pre-block frame when available recovered 24--30\% of the full-history advantage over actions alone at those same contacts. The image raised success by 7.5--8.5 percentage points (pp) per handoff, translating to 1.4--2.2\,pp per team departure because a packet can help only when the lead reaches the receiver (17.6\% of SAGE departures). Most additional successes in designed interiors came from receivers that started near the goal, whereas on human-instructed RxR routes most required travel into the goal region.

Our contributions are
\begin{enumerate}
\item A paired handoff protocol that restores the same contact state for every packet, counts missed contacts in team success, and reproduced the archived outcomes on all 6,330 same-batch baseline reruns.
\item An empirical full-history ceiling, Solo, that quantifies how much of the actions-only receiver's success deficit one earlier image recovers at a required handoff. Paired outcomes also identify the routes an image rescues or loses over an instruction-progress cue.
\item Evidence on which frame to send. In both cohorts the view before the lead's last action block outperformed the contact frame. In designed interiors an image mainly helped receivers retain a position near the goal. On human-instructed routes most additional successes required travel into the goal region.
\item A byte-priced evaluation of seven encodings with measured delivery costs. Comparing a learned sender with random senders at matched spending shows how a preference for inexpensive images can resemble a preference for informative handoffs.
\end{enumerate}
\fi

\ifextended
{\RaggedRight\section{Related Work}
\subsection{Visual history and progress in navigation}
Room-to-Room grounds natural-language routes in panoramic scans~\cite{r2r,mp3d}. Continuous VLN brings low-level motion and collisions into the task~\cite{vlnce,habitat}. RxR adds temporally aligned descriptions of guide trajectories~\cite{rxr}, and FGR2R connects sub-instructions to portions of a reference path~\cite{fgr2r}. These annotations let us locate instruction progress independently of navigation success. History-aware transformers encode prior observations~\cite{hamt}; DUET and ETPNav organize experience into maps for planning~\cite{duet,etpnav}. NaVid and StreamVLN use egocentric video context to generate navigation actions~\cite{navid,stream}. Route2Step estimates an explicit sub-instruction from the instruction and observed history~\cite{route2step}. A handoff changes whose history is available. The receiver sees the current corridor itself. The earlier image comes from the robot that traversed the preceding segment.

\subsection{Communication for cooperative perception and navigation}
Who2com and When2com learn to select cooperating agents or communication groups~\cite{who2com,when2com}. V2VNet fuses messages for joint perception and prediction, while Where2comm selects informative spatial regions under communication budgets~\cite{v2vnet,where2comm}. In collaborative VLN, Co-VLN measures the value of observations shared by agents navigating their own routes~\cite{co}. CoNavBench includes collaborative long-horizon tasks~\cite{conav}, and DeCoNav uses event-triggered dialogue to coordinate robots in its own synchronized benchmark~\cite{deco}. Mathew et al. plan truck and drone delivery routes in simulation~\cite{mathew}. The euROBIN team demonstrates physical parcel handoffs between robots~\cite{parcel}. Here the robots share one instruction. Replaying alternative packets at the same contact state lets us observe exactly which routes change outcome.

\subsection{Learning when to communicate}
CommNet learns continuous communication jointly with cooperative behavior~\cite{commnet}. TarMAC learns whom to address~\cite{tarmac}. IC3Net and attentional communication learn when to communicate or form a group~\cite{ic3net,atoc}. Early to Share, Late to Save studies event-based gating in cooperative navigation~\cite{early}. An event budget counts a detailed JPEG and a simple JPEG as one transmission each. Under a byte budget the cheaper image leaves capacity for other handoffs. Our random senders match a frozen selector's expenditure globally and within packet-size bins, so its choice of cheap images can be separated from its choice of particular routes.

\subsection{Task-oriented cost and paired evaluation}
Task-oriented communication evaluates a representation by what the receiver can accomplish with it~\cite{gunduz}. D-Lite measures navigation cost after compressing scene graphs~\cite{dlite}, and LAMP uses compact language-conditioned maps for navigation~\cite{lamp}. We charge the complete encoded application packet and score the receiver's subsequent trajectory, including path efficiency~\cite{embnav}. Reusing the same simulated state and episode seed relates our design to common random numbers in simulation~\cite{crn}. Retaining the lead run as well identifies which departures ever give a message an opportunity to help.

\par}
\fi

\section{A Paired Handoff Protocol}
\subsection{One lead trajectory, several continuations}
On each instructed route, a reference traversal yields an ordered graph of junctions and YOLO-World visual landmarks~\cite{yolo}. We keep routes with at least six nodes and choose the middle-index node for placement. A controller follows the prefix and records the reachable pose, heading, and camera view. This pose can differ from the nominal landmark.

\ifextended
In a \emph{controlled} handoff, every branch starts at this pose with the same reference-driven lead history and receives a fresh allowance of 500 continuation actions. In the \emph{autonomous} protocol, the lead runs the frozen navigator on the original instruction and its own observations. The first time it passes within 0.5\,m geodesic of the waiting receiver, the lead's history becomes available for transmission. Contact after $t$ actions leaves $500-t$ for every continuation. A lead that terminates before contact incurs zero team success and zero transmitted bytes. The 500-action team budget applies to the autonomous protocol.
On 867 missed-contact departures the simulated lead itself ended within 3\,m of the goal (354 of 789 in RxR and 513 of 2,306 in SAGE). These arrivals count as successes in the offline Solo reference over all departures. Since the required transfer never took place, each scores zero for the team and falls outside the contact-only comparisons.

Both robots run the published StreamVLN checkpoint~\cite{stream} with greedy decoding and frozen weights. The policy can stop, move forward by 0.25\,m, or turn left or right by $15^\circ$. The camera supplies $640\times480$ RGB with a $79^\circ$ horizontal field of view. For each route all receiver-packet branches share the instruction, current receiver image, full state, episode seed, and residual budget. A continuation succeeds when its final geodesic distance to the goal is at most 3\,m, at either a stop or budget exhaustion. The receiver has no goal coordinate.

\fi

Across 6,330 baseline continuations, every replay reproduced the archived success outcome, full position trajectory, terminal geodesic distance or unreachable state, and step count\ifextended{} (Appendix~\ref{app:registration})\fi.

\ifextended\section{Messages, Images and a Budgeted Sender}\fi
\subsection{Actions, progress and a prior image}
The action packet \arm{A} contains up to the last four executed lead actions in their original order. The counterfactual Count message identifies only how many actions were taken. \arm{A+I} pairs \arm{A} with the camera view immediately \emph{before} the first action in the block (\fig{f:handoff}). With a nonempty action history, the no-image branches (including A and P) copy the receiver's current view into the historical image slot. An image packet replaces that copy with the transmitted view. The receiver retains its current view in every branch. Appendix~\ref{app:registration} documents differences between packet formats, including empty histories.

The semantic packet \arm{P} adds an estimated clause index to \arm{A}. A frozen Qwen2.5 text model~\cite{qwen} first divides the shared instruction into contiguous clauses. The frozen Route2Step aligner~\cite{route2step} then uses only the lead's observed history. A unique phrase match identifies a clause. Ambiguous estimates produce an explicit unknown. The receiver expands that code into a sentence about unknown progress. \arm{P+I} and \arm{P+T} append an image to \arm{P} and preserve the estimate, including unknowns.

For a transmitted pre-block image, \arm{A+I} and \arm{P+I} use the same navigator-preprocessed $384\times384$ JPEG with different headers. \arm{P+T} contains a $320\times240$ thumbnail resized directly from the same camera frame. Both encodings use JPEG quality 90. All 492 autonomous contacts in SAGE have four actions, giving a 6-byte \arm{A} packet and an 8-byte \arm{P} packet. We measure complete application payloads after serialization. We measure application delivery separately on shaped links and a workstation--server connection (Appendix~\ref{app:links}).

\subsection{A sender with a byte budget}
For contact $i$ let $Y_i^{P+I},Y_i^{P}$ be the two observed continuation outcomes and let $c_i=B_i^{P+I}-B_i^{P}$ be the additional bytes for the model-view image. We fit a ridge model on development contacts to predict $Y_i^{P+I}-Y_i^{P}$. Its inputs are instruction-clause entropy given the actions, entropy from CLIP image--text similarities~\cite{clip}, the entropy reduction from combining both distributions, and log incremental bytes. The sender transmits for positive predicted gains per byte above a development-set threshold. The threshold targets the development cost of always sending a thumbnail. Features, coefficients, bin edges, and the threshold are frozen before scene transfer.

Let $z_i$ be the selector's binary send decision. A random sender can match its expenditure by transmitting at probability
\begin{equation}
 p=\frac{\sum_i z_i c_i}{\sum_i c_i}.\label{eq:spend}
\end{equation}
\ifextended
Its expected route outcome is $Y_i^{P}+p(Y_i^{P+I}-Y_i^{P})$. A second random sender computes $p_b$ by restricting both sums to the same packet-size bin $b$. Both policies buy the same incremental bytes in expectation. Only the second also matches spending within each cost stratum. We calculate their expected outcomes from the measured paired continuations, including the shared short-packet cost on contact. The alternatives can thus be compared without Monte Carlo noise from an additional sampled transmission schedule.

\fi

\section{Experimental Setup}
We first developed the protocol on 150 controlled R2R handoffs in 11 scanned Matterport3D buildings~\cite{r2r,mp3d}. We then fixed 1,122 English human-instructed RxR routes~\cite{rxr} in the same buildings, excluding geometry used in development in either direction. This is a new-route test of the action/image comparison. Contacts in these RxR buildings subsequently supplied development data for the sender and cost selector. Calibration of progress estimates used separate R2R training prefixes.

For new-scene transfer we use Area-based English instructions from SAGE-Bench and artist-designed InteriorGS environments~\cite{sage,interiorgs}. Habitat supplies collision-aware movement~\cite{habitat}, and the scenes' Gaussian splats supply RGB. The formal scenes come from the dataset's published training directory and were held out from our local design decisions. We froze eligible routes before running navigation: 95 scenes supplied 2,798 departures (\tab{t:cohorts}).
\cohortstable
For condition $k$ let $Y_{si}^k$ denote team success on departure $i$ in scene $s$, including zero on a missed contact. For the original SAGE comparisons, our primary estimate of the success increment from condition $j$ to $k$ is
\begin{equation}
 \widehat{\Delta}_{k,j}=\frac{100}{K}\sum_{s=1}^{K}\frac{1}{n_s}
                     \sum_{i=1}^{n_s}(Y_{si}^k-Y_{si}^j), \label{eq:delta}
\end{equation}
where $K=95$ and $n_s$ is the number of departures in that scene. This scene-equal estimate averages the 95 scene-specific success differences. Within a scene, the effect per departure equals its contact rate times its effect per contact, with a zero contribution for a scene with no contact. RxR uses pooled paired differences over the 1,122 routes in its 11 fixed buildings.

The original five SAGE comparisons use scene-equal team success over all 2,798 departures (Table~\ref{t:cohorts}(b)). We bootstrap whole scenes 10,000 times, preserve route pairings and refit the matched send probabilities in each draw. Two-sided 99\% percentile intervals give a Bonferroni-adjusted family confidence level of at least 95\% under the bootstrap approximation. The original RxR analysis instead resamples routes within each of the 11 fixed buildings and uses 97.5\% intervals for its two planned contrasts. Differences in success rates are given in percentage points (pp). We specified no noninferiority margin for the thumbnail.

\ifextended
The four revision comparisons (N1--N4) form a separate prespecified family. Their question concerns what a receiver does with a packet, so the primary estimate pools paired routes at realized handoffs (N1--N2) or controlled placements (N3--N4). Each uses a two-sided 98.75\% interval, resampling whole scenes in SAGE or paired routes within buildings in RxR. Table~\ref{t:cohorts}(b) identifies the primary estimate and denominator for each contrast. For autonomous image comparisons we also give the complementary estimate across departures or among contacts. A controlled placement has no observed autonomous contact rate from which to estimate a team effect. The per-contact intervals for the original image comparisons were planned secondary analyses (T1.2). Other secondary intervals are pointwise 95\% unless specified.

\textit{Experimental scale.}
The 106 formal environments cover 3,920 autonomous departures and controlled placements on the same 3,920 routes. The original and registered revision batches executed 28,829 receiver continuations (Table~\ref{t:cohorts}), including the 6,330 same-batch reruns. These batches used 224.59 GPU-worker hours, including preparation.
S1 separately continued the lead at \SOneNewSoloRuns{} contacts and reused terminal logs for the other 3,095 departures. Including its replay check, S1 used \SOneSoloGPUHours{} additional GPU-worker hours. The combined allocation for these formal batches and S1 was 230.40 GPU-worker hours. The per-departure records accompany the paper.

\fi

\section{Results}
\subsection{One image changes outcomes in both directions}\label{sec:oneimage}
One pre-block image raised success at realized handoffs by 7.5--9.4\,pp and reduced mean terminal error by 0.5--1.2\,m across the action and estimated-progress comparisons (Table~\ref{t:receiver_navigation}). Over executed actions, the per-handoff gain was $7.51\pp$ in RxR and $8.54\pp$ in SAGE. These correspond to relative increases of 17.4\% and 21.8\%.
\effectsfig

The corresponding gains in team success were $2.23\pp$ on RxR and $1.44\pp$ on SAGE. In RxR the image rescued 57 routes and reversed 32 successes. Figure~\ref{f:effects} gives the adjusted intervals, the count-only action comparison and the results under controlled placement.

In SAGE the estimated-progress packet already identified a clause when the sender could resolve it. Adding the image increased success by $9.35\pp$ per realized handoff (planned secondary 99\% CI $[+2.25,+16.29]$), a 23.2\% relative increase. Across all departures the scene-equal team gain was $1.64\pp$.
 
That SAGE increment comprised 100 previously unsuccessful routes rescued and 54 successes reversed. The image added 26.81\,kB per contact on average. Table~\ref{t:cost} gives the full application cost per departure, charging zero on missed contacts.
\costtable
\ifextended\begin{table*}[t]
\caption{Receiver navigation among realized autonomous handoffs. Differences are paired against the indicated base packet.}
\label{t:receiver_navigation}
\centering\fontsize{8.3}{10}\selectfont\setlength{\tabcolsep}{3pt}
\begin{tabular*}{\textwidth}{@{\extracolsep{\fill}}llrrrrrrrr@{}}\toprule
Cohort & Packet (base) & Contacts & SR (\%) & $\Delta$SR (pp) & \shortstack[r]{Median NE\\(m)} & \shortstack[r]{$\Delta$ mean NE (m)\\{[95\% CI]}} & \shortstack[r]{Outside OSR\\(\%)} & SPL & \shortstack[r]{Median\\actions} \\ \midrule
RxR & \textsf{Count} (base) & 333 & 39.64 & --- & 4.95 & --- & 48.20 & 0.291 & 72 \\
RxR & \textsf{A} (Count) & 333 & 43.24 & $+3.60$ & 3.82 & $-0.40$ [$-1.30,+0.47$] & 56.31 & 0.307 & 76 \\
RxR & \textsf{A+I} (A) & 333 & 50.75 & $+7.51$ & 2.36 & $-1.17$ [$-2.00,-0.34$] & 65.32 & 0.414 & 62 \\
RxR & \textsf{Donor} (A) & 321 & 45.79 & $+3.43$ & 3.25 & $-0.95$ [$-1.75,-0.15$] & 58.18 & 0.347 & 63 \\
RxR & \textsf{At contact} (A+I) & 321 & 43.93 & $-6.23$ & 3.38 & $+0.55$ [$-0.26,+1.35$] & 59.55 & 0.318 & 68 \\
SAGE & \textsf{A} (base) & 492 & 39.23 & --- & 3.79 & --- & 65.49 & 0.244 & 38 \\
SAGE & \textsf{A+I} (A) & 492 & 47.76 & $+8.54$ & 3.10 & $-0.46$ [$-0.74,-0.16$] & 65.49 & 0.311 & 32 \\
SAGE & \textsf{P} (base) & 492 & 40.24 & --- & 3.77 & --- & 64.79 & 0.252 & 38 \\
SAGE & \textsf{P+I} (P) & 492 & 49.59 & $+9.35$ & 3.04 & $-0.50$ [$-0.80,-0.19$] & 66.20 & 0.325 & 32 \\
SAGE & \textsf{P+T} (P) & 492 & 48.78 & $+8.54$ & 3.07 & $-0.46$ [$-0.75,-0.17$] & 64.79 & 0.321 & 32 \\
\bottomrule\end{tabular*}
\par\smallskip\parbox{0.96\textwidth}{\footnotesize Finite distances and SPL are available for 318 of 333 RxR contacts and all 492 SAGE contacts. Outside OSR measures goal-region entry for receivers starting outside (222 RxR and 142 SAGE). The two RxR substitution rows use 321 nonempty handoffs and their own controls (A 42.37\%, A+I 50.16\%). Distances and SPL cover 310 of these handoffs, outside OSR covers 220.}
\end{table*}
\fi

\subsection{One frame relative to the full-history ceiling}
The full-history Solo ceiling exceeded actions-only receiver success by $35.37\pp$ per realized SAGE handoff (pointwise 95\% CI $[29.22,41.26]$) and $24.92\pp$ per realized RxR handoff ($[18.91,30.75]$). In this post-registration supplementary comparison Solo retains the lead's own pose, heading, visual history and pending actions. The receiver starts at its nearby pose with the same remaining budget. The upper panel of Table~\ref{t:solo_s1} and \fig{f:geometry}(c) use the same contacts in each cohort.
\SOneSoloTable

Sending one earlier frame when available, from a median history of 41 frames (IQR 23--59 across the 825 contacts), recovered 24--30\% of the full-history advantage over actions alone. History counts the lead's RGB observations before it processes the contact view. The median was 40 frames in SAGE (IQR 23--57) and 43 in RxR (IQR 26--62). Twelve RxR routes reached the receiver at the start. Their histories contain zero frames, both packet arms use the same input on those routes, and all twelve contribute to the 333-contact estimate.

In SAGE the image added 42 net successes against a 174-success gap between Solo and actions alone. It recovered 24.14\% of that gap (paired 95\% CI 11.25--35.26\%). In RxR the corresponding counts were 25 and 83, giving 30.12\% recovery (9.09--50.00\%). These percentages compare net changes across the same routes.

In the SAGE comparison of Solo with the progress-and-image receiver, both succeeded on 206 routes. The receiver alone succeeded on 38, the lead alone on 161, and neither on 87. The lower panel of Table~\ref{t:solo_s1} separately uses every departure for every condition. It covers 2,798 in SAGE with equal weight per scene and 1,122 in RxR with routes pooled.

\subsection{Contact opportunity and distance to the goal}\label{sec:opportunity}
Where the gains occur depends on whether the receiver still has to travel. In a post hoc stratification by actual starting pose, 40 of the 46 net SAGE gains over the progress cue came from receivers already inside the scoring radius. With actions as the common reference, most SAGE net gains still came from inside, while 22 of 25 RxR net gains came from outside.

Only 492 of 2,798 SAGE departures reached contact (\fig{f:geometry}). Of these, 350 receivers started within the goal radius. Figure~\ref{f:geometry}(c) gives the immediate-stop reference per realized handoff. Table~\ref{t:cost} reports it separately per team departure. The projected landmark and the actual reachable pose can lie at different distances from the goal (Appendix~\ref{app:sender}).
\geometryfig

\subsection{How many bytes does the image need?}
\casesfig
\ifextended At about one quarter of the full payload, the SAGE image still delivered most of the observed net increment over the progress packet. Gains were detectable from about 7\,kB per handoff. None was detected at 3.7\,kB or below (\fig{f:e2_rate}, pointwise 95\% intervals).

\begin{figure}[t]
\centering
\includegraphics[width=\columnwidth]{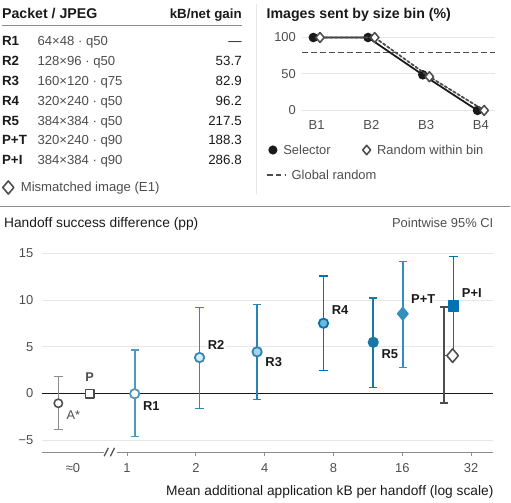}
\caption{Value of seven encodings of the same lead frame. Each blue point adds an image to the same progress packet P at 492 SAGE handoffs. Whiskers give pointwise 95\% intervals from resampling whole scenes. The open diamond gives the mismatched image versus its E1 progress control, with its whisker offset to the left for visibility. The near-zero points are the 8-byte reference P and the 6-byte actions-only packet A* compared with its archived P. The key gives decimal kB per net additional success (Table~\ref{t:e2_all}). The inset shows the fraction of images sent across ascending size bins (231, 150, 86 and 25 contacts). Bin edges fixed on development data were 25,938, 32,496 and 40,801 additional bytes. Both random policies match the selector in expected spending.}\label{f:e2_rate}
\end{figure}

At 7.23\,kB the image added $7.52\pp$ per handoff, compared with $9.35\pp$ for the 26.81\,kB full image. Table~\ref{t:e2_all} gives the interval, successful routes gained and lost, and byte efficiency at every encoding. The curve reports pointwise intervals. The planned secondary 99\% interval for the original full-image comparison appears in Section~\ref{sec:oneimage}.
\fi
The fixed quality-90 thumbnail retained 42 net additional successes, compared with 46 for the full image, with 40\% fewer application bytes. Their paired team difference was $-0.25\pp$ (pointwise 95\% CI $[-0.95,+0.42]$).

On the emulated UDP link, accounting for measured delivery rates gives an expected gain per handoff of $6.84\pp$ for the 7\,kB image and $6.03\pp$ for the full image if a lost image leaves the short packet available (Table~\ref{t:e2_all}).

\subsection{Can the sender choose which handoffs need an image?}
Matching expected spending within packet-size bins removed the learned sender's apparent advantage. The sender exceeded globally random transmission by $0.32\pp$ in team success (99\% CI $[-0.11,+0.78]$). At the same expenditure within bins its difference from random was $-0.01\pp$ (95\% CI $[-0.16,+0.14]$). A cost-stratified random baseline measures what a sender adds beyond buying less expensive pictures (\fig{f:perturb}).

The selector sent every image in the two cheapest bins and none in the most expensive (\fig{f:e2_rate}). It sent 423 images in total, compared with 420.70 expected under size-matched random transmission. Table~\ref{t:cost} gives the other operating points. The full image cost about 287\,kB per additional success over the short packet, compared with 188\,kB for the fixed thumbnail.

\ifextended\subsection{Which frame should the lead send?}
Sending the pre-block frame instead of the contact frame raised success by 6--7\,pp per handoff in both cohorts. The paired secondary differences were $+6.91\pp$ in SAGE and $+6.23\pp$ in RxR. Their pointwise 95\% intervals excluded zero (\fig{f:perturb}). The contact frame was modestly more similar to the receiver's current view in both cohorts (Appendix~\ref{app:substitution}).

Moving the SAGE frame eight actions earlier gave $-1.22\pp$ per handoff (95\% CI $[-5.96,+3.39]$) relative to the pre-block frame. Table~\ref{t:substitution} lists the receiver outcomes for each frame time.

\begin{figure}[t]
\centering
\includegraphics[width=\columnwidth]{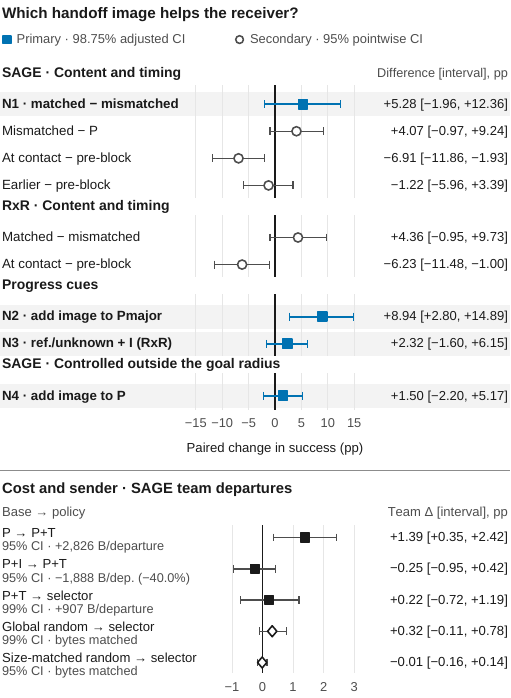}
\caption{Paired changes in receiver success. Content and timing use 492 autonomous SAGE handoffs and 321 autonomous RxR handoffs with nonempty histories. Shaded rows with filled squares show the four registered comparisons with 98.75\% intervals. N1 tests route matching and N2 adds an image to a constant first-clause index in autonomous SAGE. N3 adds an image to the reference cue at 1,122 controlled RxR placements (1,074 unknown indices). N4 adds it to estimated progress at 1,136 controlled SAGE placements outside 3\,m. Open circles show secondary, pointwise 95\% intervals. SAGE resamples scenes and RxR resamples routes within each building. The separate cost and sender panel shows scene-equal differences in team success per SAGE departure, the original 95\% or 99\% intervals, and differences in application bytes per departure. Open diamonds mark randomized-policy expectations. Table~\ref{t:cost} gives absolute outcomes.}\label{f:perturb}
\end{figure}

The matched pre-block frame gave 244 successes against 218 for the donor in SAGE, but N1 did not resolve an exact-route advantage ($+5.28\pp$, 98.75\% CI $[-1.96,+12.36]$). Most donors preserved the scene (\fig{f:gain_gradient}). Appendix~\ref{app:substitution} gives the RxR comparison and the exploratory two-cohort estimate.

\begin{figure*}[t]
\centering\includegraphics[width=\textwidth]{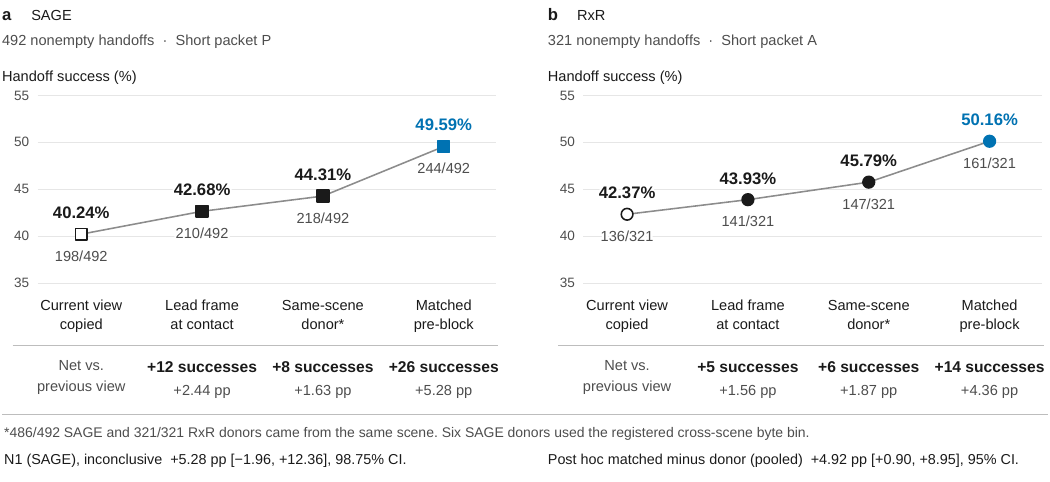}
\caption{Gain decomposition across four handoff views. Success is measured on the same 492 SAGE contacts and 321 nonempty RxR contacts within each panel. The first condition copies the receiver's current view into the history slot. The others replace that copy with the lead's contact frame, a mismatched donor frame, or the matched pre-block frame. Donors came from the same scene on 486/492 SAGE and all 321 RxR handoffs. The six other SAGE donors met the registered byte-bin rule. The authors chose the display order. Adjacent differences are descriptive paired contrasts. For the final contrast, the registered SAGE N1 interval includes zero. The two-cohort estimate is post hoc.}\label{f:gain_gradient}
\end{figure*}
\subsection{Image value with alternative progress cues}
With a constant first-clause index, adding the image raised SAGE handoff success by $8.94\pp$ (N2, 98.75\% CI $[+2.80,+14.89]$). This 8-byte progress packet specifies the same clause on every route. Table~\ref{t:five_packets} places the success counts beside those for actions and estimated progress.

The index matched 169 of 171 scorable R2R training prefixes aligned to FGR2R sub-instructions~\cite{fgr2r}. The route-dependent aligner matched 138. These are calibration results. Across all SAGE departures the scene-equal team gain with the constant index was $1.60\pp$ (secondary 98.75\% CI $[+0.50,+2.72]$). Appendix~\ref{app:registration} gives the paired changes in navigation error and SPL.

\begin{table}[t]
\caption{Successes on the same 492 autonomous SAGE handoffs. A carries executed actions. P adds estimated progress, and $P_{\rm major}$ adds a constant first-clause index to A. I adds the pre-block image.}\label{t:five_packets}
\centering\small
\begin{tabular*}{\columnwidth}{@{\extracolsep{\fill}}lrrrrr@{}}\toprule
Packet & A & P & $P_{\rm major}$ & P+I & $P_{\rm major}{+}I$ \\
\midrule
Successes & 193 & 198 & 201 & 244 & 245 \\
Success (\%) & 39.23 & 40.24 & 40.85 & 49.59 & 49.80 \\
\bottomrule\end{tabular*}
\par\smallskip\parbox{\columnwidth}{\footnotesize A uses the archived actions-only run. The other four counts use the E3a batch. Counts across columns are descriptive.}
\end{table}

At controlled RxR placements with a mostly unknown reference index, N3 did not resolve the image increment ($+2.32\pp$, 98.75\% CI $[-1.60,+6.15]$). The two no-image packet formats changed 72 of 1,122 outcomes with zero net change in successes (Appendix~\ref{app:registration}).
\subsection{Holding a position near the goal and reaching it from outside}
The pre-block image reduced SAGE walkouts by $11.43\pp$ among receivers that started within the goal radius (post hoc, 95\% CI $[-17.97,-4.99]$). Walkouts fell from 63.1\% to 51.7\% (\fig{f:receiver_behaviour}). Neither arm stopped on the first action in these handoffs.

Among RxR receivers starting outside the goal radius, adding the image to actions raised the probability of entering it by $9.01\pp$ (95\% CI $[+2.25,+15.77]$). Entry rose from 56.3\% to 65.3\%.

\begin{figure}[t]
\centering\includegraphics[width=\columnwidth]{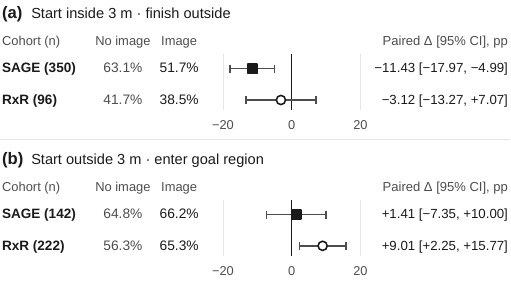}
\caption{Receiver behavior at autonomous handoffs with known starting distance (post hoc). (a) Fraction finishing outside after starting within 3\,m. (b) Fraction ever entering the goal region after starting outside. Columns compare P with P+I in SAGE and A with A+I in RxR. Whiskers give paired, pointwise 95\% intervals for the differences in percentage points.}\label{f:receiver_behaviour}
\end{figure}

In controlled SAGE placements outside the radius, N4 did not resolve an image gain over estimated progress ($+1.50\pp$, 98.75\% CI $[-2.20,+5.17]$), and the earlier actions-only comparison went from 385 to 369 successes ($-1.41\pp$, pointwise 95\% CI $[-4.44,+1.56]$).
\fi

\subsection{Paired continuations with opposite outcomes}
In \fig{f:cases}(a), the receiver with the short packet first turns right. The receiver with the added image initially advances and finishes within a metre of the goal. In \fig{f:cases}(b), adding the same kind of image turns a success into a failure. The image-conditioned receiver ends 11.27\,m from the goal. The gain case starts 5.15\,m from the goal along the navigation mesh. In the loss case, the goal lies just across a wall. The two continuations terminate on opposite sides of the 3\,m scoring boundary.\ifextended{} The corresponding first-person views appear in \fig{f:views}.\fi

\section{Discussion}
\textit{Design implications.}
On the same 492 SAGE handoffs, success was 39.2\% with actions alone, 47.8\% with actions plus the pre-block image, 71.1\% with an immediate stop, and 74.6\% at the Solo ceiling (\fig{f:geometry}(c)). Stopping scores well here because 350 receivers already stood within the goal radius at contact. For receivers that continued from those near-goal starts, the trajectories show how the image helped (post hoc). Adding it to the estimated-progress packet reduced the fraction finishing outside by $11.43\pp$.

On the 333 RxR handoffs, actions alone scored 43.2\%. Only 96 receivers started inside the goal radius, so immediate stopping scored 28.8\%, below the 50.8\% achieved with actions plus an image. The full-history Solo ceiling was 68.2\% on those same handoffs. Among the receiver choices in \fig{f:geometry}(c), adding the earlier image was the only change that improved on actions alone in both cohorts. A SAGE encoding of about 7\,kB retained much of this gain.

\textit{Evaluation implications.}
Among nonempty RxR prefixes with unknown progress, two no-image prompts differing by seven reconstructed tokens changed nearly 6\% of outcomes. In SAGE the lowest-rate and full images each flipped roughly three handoffs in ten. The lowest-rate image produced no net gain, while the matched full image yielded 46 net successes. Pairing reveals both the frequency and direction of these changes (\fig{f:gross_flips}). For senders operating under a byte budget, matching random transmission within size bins separates route selection from a preference for inexpensive frames.

\textit{Limitations.}
The simulations use one frozen navigator, one waiting receiver and one opportunity to exchange a packet. Solo provides an empirical ceiling for this navigator on these handoffs. The lead's own contact pose, heading and pending actions accompany its history. We did not test packets carrying multiple historical frames. They are the natural next step between transmitting one frame and retaining the lead's full history. For context, the updated RGB-only StreamVLN achieved 56.4\% success on standard R2R-CE Val-Unseen and 54.4\% on RxR-CE (\href{https://arxiv.org/html/2507.05240v2#S3.T1}{Table I} of~\cite{stream}). Our receiver takes over mid-instruction, team success requires contact, and SAGE changes the visual domain. Layout, language and geometry vary together across cohorts. Packet format can also change the prompt, and the thumbnail and model-view image use different preprocessing paths. The registered comparisons did not resolve exact-route specificity or image gains at controlled RxR placements with mostly unknown progress and controlled SAGE placements outside the goal region. Appendix~\ref{app:links} gives delivery costs measured separately from navigation.

\section{Conclusion}
\ifextended
One image from before the lead's last action block raised receiver success at realized handoffs in both scanned buildings and held-out designed interiors. Relative to the empirical full-history ceiling, this frame recovered about a quarter of the actions-only deficit in SAGE and three tenths in RxR. The earlier frame outperformed the view at contact, and much of the SAGE gain survived compression to about 7\,kB. In designed interiors the image mainly helped receivers stay near the goal. On human-instructed routes it helped them reach it. Same-state replay measured these changes route by route. Matching spending within packet-size bins identified how much of a sender's apparent advantage came from choosing cheaper images.
\fi

\ifextended\useRomanappendicesfalse
\appendices
\setcounter{figure}{0}\setcounter{table}{0}
\renewcommand{\thefigure}{A\arabic{figure}}
\renewcommand{\thetable}{A\arabic{table}}
\renewcommand{\theHfigure}{supplement.\arabic{figure}}
\renewcommand{\theHtable}{supplement.\arabic{table}}

\section{Route Construction and Paired Outcomes}\label{app:packet}
\textit{Placement and replay.}
Reference traversal defines the ordered landmark graph. The placement controller takes at most 400 forced actions along the prefix and records its final reachable pose and current view. The pose includes the full camera rotation. In the designed scenes the camera sits 1.25\,m above the agent base. The agent radius is 0.10\,m and its height is 1.50\,m. We cast vertically onto the navigable surface on the start island to recover the floor height for the horizontal path. Unavailable floor projections cause a geometry exclusion before evaluation.

\textit{Frozen route sets.}
The English RxR Guide source yields 1,223 unique routes. After removing 31 routes shared with prior experiments, 46 internal geometric aliases, and 24 routes that fail the six-node rule, the cohort contains 375 US-English and 747 Indian-English instructions. One original English instruction per route was selected before qualification. \tab{t:rxrcounts} reports the successes, episodes gained, episodes lost and ties for each condition.

\begin{table}[!htbp]
\caption{Per-condition success counts and outcomes relative to the preceding message}\label{t:rxrcounts}
\centering\tfont\setlength{\tabcolsep}{2.5pt}
\begin{tabular*}{\columnwidth}{@{\extracolsep{\fill}}lrrrr@{}}
\toprule
Message & {Successes} & {Gained} & {Lost} & {Tied}\\
\midrule
\multicolumn{5}{@{}l}{\emph{Protocol development (R2R, controlled, 150 placements)}}\\
None &34&{---}&{---}&{---}\\
Count &34&24&24&102\\
\arm{A} &42&26&18&106\\
\arm{A+I} &48&17&11&122\\
\addlinespace[2pt]
\multicolumn{5}{@{}l}{\emph{RxR, autonomous lead, 1,122 departures}}\\
Count &132 &{---}&{---}&{---}\\
\arm{A} &144&55&43&1024\\
\arm{A+I} &169&57&32&1033\\
\addlinespace[2pt]
\multicolumn{5}{@{}l}{\emph{RxR, controlled placement, 1,122 continuations}}\\
Count &331&{---}&{---}&{---}\\
\arm{A} &355&169&145&808\\
\arm{A+I} &395&173&133&816\\
\bottomrule\end{tabular*}
\tnote{Ties include joint failures. R2R development intervals for actions over Count and image over actions are $[-3.38,+14.00]$ and $[-2.78,+10.71]$ pp (pointwise 95\%). RxR uses separate formal intervals.}
\end{table}

For SAGE we take the first published Area-based instruction for each candidate and deduplicate reversed anchor/geometry pairs. The horizontal endpoints must be more than 3\,m apart. A fixed scene-specific shuffle determines the order of candidates, with at most 30 qualified routes per scene. Of 3,845 examined candidates, 947 fail geometry qualification and 100 have too few nodes. Object-carrying instructions are evaluated on travel to the destination. The navigation action set provides no manipulation action.

\textit{Scene variation and contact geometry.}
\fig{f:scenes} weights each scene once regardless of how often contact occurs. Leaving any single scene out gives a scene-mean \arm{P}-to-\arm{P+I} effect between $+1.44$ and $+1.76\pp$. On the 492 departures with contact, the two messages both succeed on 144 and both fail on 194. Adding the image rescues 100 of the 294 short-packet failures and loses 54 of the 198 short-packet successes.
\begin{figure*}[t]
\centering
\includegraphics[width=\textwidth]{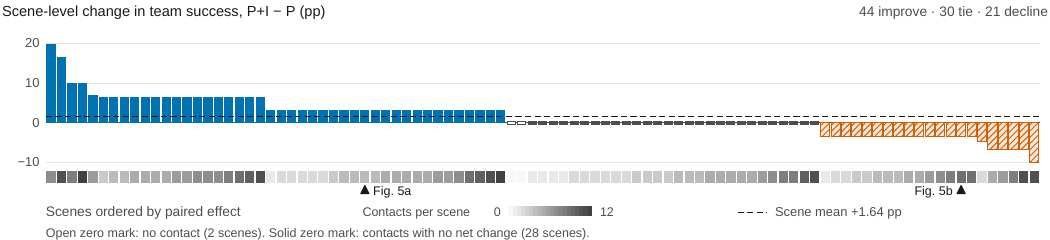}
\caption{Per-scene paired effect in SAGE. Each bar gives the difference in team success between \arm{P+I} and \arm{P} over all departures in one scene. The gray band gives the number of contacts. Two scenes had no contact, so every packet had the same outcome by design. Triangles locate the two cases in Fig.~\ref{f:cases}. Every scene enters the scene-equal mean in~(\ref{eq:delta}).}
\label{f:scenes}
\end{figure*}

The nominal graph point can lie at the center of an object, making its geodesic undefined. The actual receiver pose has a finite geodesic on all 2,798 SAGE departures. In \fig{f:nominal}, 44 departures with a nominal point inside the scoring radius actually start outside, and 86 switch categories in the opposite direction. \tab{t:actual} counts outcomes by actual start. \tab{t:nominal} instead uses the design-time point and the online clause estimate. In SAGE, all contacted receivers had at least 123 actions left, so an exhausted budget does not account for their starting-to-terminal difference.
\begin{figure}[t]
\centering\includegraphics[width=\columnwidth]{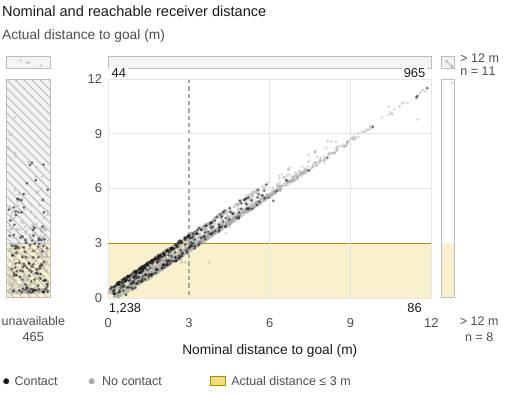}
\caption{Nominal and reachable receiver distance on 2,798 SAGE departures. Both axes give geodesic distance to the goal. The four corner counts include all finite pairs, including observations above 12\,m in the marginal strips. The left strip preserves the actual distances of all 465 departures with unavailable nominal distance. Of these, 338 are within the scoring radius and 127 are outside. The yellow band marks an actual distance of at most 3\,m.}
\label{f:nominal}
\end{figure}

\begin{table}[t]
\caption{Outcomes by actual start distance to goal, post hoc}\label{t:actual}
\centering\tfont\setlength{\tabcolsep}{2pt}
\begin{tabular*}{\columnwidth}{@{\extracolsep{\fill}}l S[table-format=4.0] S[table-format=3.0] *{4}{S[table-format=3.0]}@{}}
\toprule
& & & \multicolumn{4}{c}{Team successes}\\
\cmidrule(lr){4-7}
Actual start & {Departures} & {Contacts} & {\arm{A}} & {\arm{A+I}} & {\arm{P}} & {\arm{P+I}}\\
\midrule
\multicolumn{7}{@{}l}{\emph{RxR, autonomous lead}}\\
$\leq3$\,m &243&96&56&59&{---}&{---}\\
$>3$\,m &860&222&88&110&{---}&{---}\\
Unavailable &19&15&0&0&{---}&{---}\\
All &1122&333&144&169&{---}&{---}\\
\addlinespace[2pt]
\multicolumn{7}{@{}l}{\emph{SAGE, autonomous lead}}\\
$\leq3$\,m &1662&350&125&162&129&169\\
$>3$\,m &1136&142&68&73&69&75\\
All &2798&492&193&235&198&244\\
\bottomrule\end{tabular*}
\tnote{All SAGE start distances were available. Dashes indicate that this condition was outside the formal RxR comparison.}
\end{table}

\begin{table}[t]
\caption{Nominal position and estimated instruction-progress strata in SAGE}\label{t:nominal}
\centering\tfont\setlength{\tabcolsep}{2pt}
\begin{tabular*}{\columnwidth}{@{\extracolsep{\fill}}l S[table-format=4.0] S[table-format=3.0] *{2}{S[table-format=3.0]}@{}}
\toprule
Group &{Departures}&{Contacts}&{\arm{P} successes}&{\arm{P+I} successes}\\
\midrule
\multicolumn{5}{@{}l}{\emph{Nominal landmark-to-goal distance, all departures}}\\
$\leq3$\,m &1282&266&110&134\\
$>3$\,m &1051&100&42&42\\
Unavailable &465&126&46&68\\
\addlinespace[2pt]
\multicolumn{5}{@{}l}{\emph{Progress message, contacted routes only}}\\
Clause index given &{---}&270&117&141\\
Unknown &{---}&222&81&103\\
\bottomrule\end{tabular*}
\tnote{Both packets carry the same progress estimate on each route. Determining the clause is an online operation performed at contact.}
\end{table}

\section{Reconstructing the Message and Sender}\label{app:sender}
The R2R development counts in \tab{t:rxrcounts} describe how the interface was selected. The formal RxR episodes use new route geometries. \fig{f:views} follows the two SAGE cases from contact to termination, with the archived episode seed, receiver state, and renderer. Each pair starts with the same camera view.

\begin{figure*}[t]
\centering\includegraphics[width=\textwidth]{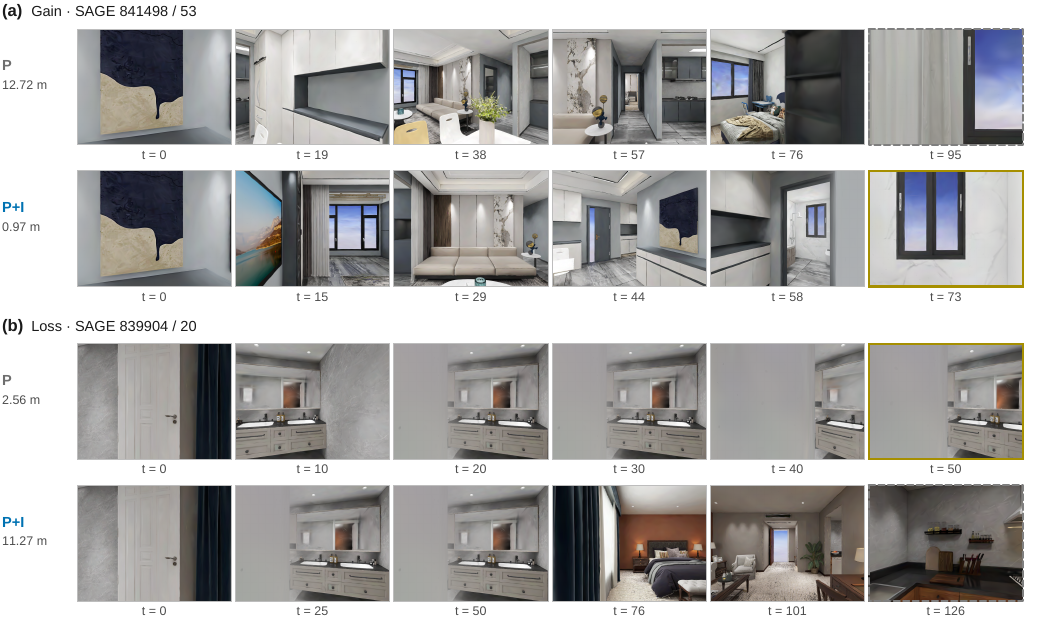}
\caption{Receiver views under \arm{P} and \arm{P+I} for the paired examples in \fig{f:cases}. Frames show contact, 20\%, 40\%, 60\%, 80\% and the final action in each recorded continuation. Each label is an executed step. Deterministic replay used the original episode seed and renderer. The four end positions and terminal goal distances matched their records. Solid gold outlines mark final views within 3\,m geodesic of the goal. Dashed gray outlines mark final views outside.}
\label{f:views}\end{figure*}

The paired radius analysis in \fig{f:radius_sensitivity} rescores the same final positions. The strata in \fig{f:handoff_strata} describe how image effects vary across observed routes.
\begin{figure}[t]
\centering\includegraphics[width=\columnwidth]{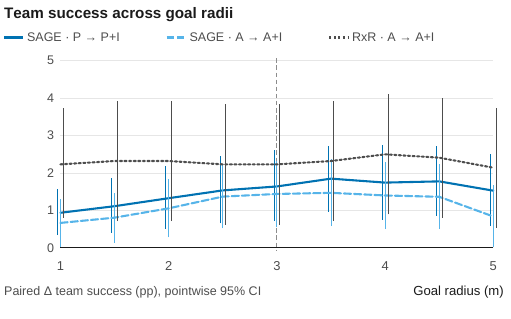}
\caption{Team success when rescoring the archived terminal positions at different goal radii. Each point is the paired image increment per autonomous departure. Whiskers give pointwise 95\% intervals. The vertical guide marks the primary 3\,m success radius.}\label{f:radius_sensitivity}
\end{figure}
\begin{figure}[t]
\centering\includegraphics[width=\columnwidth]{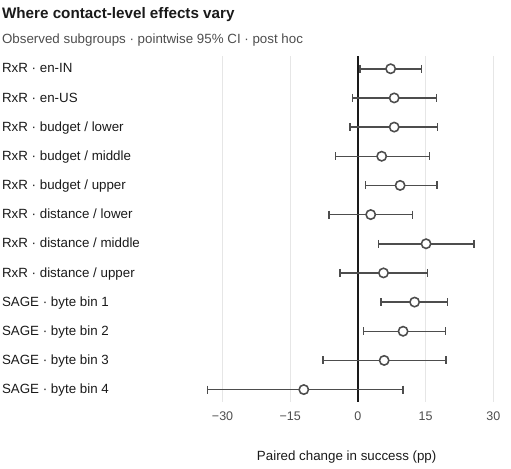}
\caption{Post hoc differences in success by RxR language, remaining actions and starting distance, and by SAGE image-byte bin. Handoffs are paired within each displayed group. Intervals are pointwise 95\%.}\label{f:handoff_strata}
\end{figure}

{\RaggedRight \textit{Serialized packet and model input.}
We evaluated the \href{https://huggingface.co/mengwei0427/StreamVLN_Video_qwen_1_5_r2r_rxr_envdrop_scalevln}{released StreamVLN checkpoint}. \arm{A} has a one-byte version, a one-byte count and one byte per action. \arm{A+I} carries length-prefixed JSON action/index metadata and a length-prefixed JPEG. \arm{P} uses a one-byte version, one-byte count, two-byte clause index and the executed actions. \arm{P+I} and \arm{P+T} extend \arm{P} by a four-byte length and the JPEG. Multibyte fields use little-endian order. \arm{P+T} uses bicubic resizing from $640\times480$ followed by one encoding. In the example in \fig{f:handoff}, A+I and P+I contain the same 24,274-byte JPEG. Their headers cost 63 and 8 bytes, respectively, plus a 4-byte image-length field for both. We charge for every byte actually serialized.\par}

The last action block may be short. Among the 333 RxR contacts, 321 have a nonempty lead history. The 12 empty histories use the same no-packet continuation in each arm. The 789 missed contacts and all 12 empty histories are retained in the team denominator. At a nonempty contact, the decoded action turn and the current RGB accompany the original instruction. No reference trajectory or goal is an online model input.

\textit{Where team outcomes changed.}
In the original SAGE comparison of estimated progress with and without the image, 44 scenes had a positive difference, 30 were tied and 21 had a negative difference. On the 1,051 departures with a finite nominal landmark distance above 3\,m, each packet scored 42 successes. Using actual contact poses, 37 of the 42 additional actions-plus-image successes started inside 3\,m. In the progress comparison, six of the 46 net additional successes began outside. RxR's corresponding actions-plus-image contrast assigned 22 of its 25 net gains to starts outside the radius. The nominal landmark and the reachable receiver pose need not coincide.

\textit{Clause reconstruction.}
The Qwen2.5-7B segmenter uses greedy decoding. The released Route2Step multimodal instruction aligner consumes the observed prefix, with thirteen earlier and three recent views under its fixed sampling rule. A normalized match to a unique contiguous phrase in a single shared clause yields an index. Index zero encodes an unknown. The calibration generator matched R2R training-route prefixes to FGR2R sub-instructions and their published viewpoint ranges~\cite{fgr2r}. Of 179 sampled prefixes, 171 had a unique label. A constant first-clause index matched 169/171 on these R2R calibration routes. Among the 100 scorable multi-clause cases, the estimator matched 81 and the constant first-clause index matched 98. Reference-route alignments do not directly label autonomous lead trajectories that depart from the reference. Both SAGE packets share each online estimate, as recorded in \tab{t:nominal}.

\textit{Selector coefficients and operating point.}
Let $a_j$ be the text model's probability of clause $j$ given the action prefix and instruction, $v_j$ the normalized image--clause similarity, and $h_j\propto a_jv_j$. The predictor inputs are
\begin{equation}
 \bigl[H(a)/L,\ H(v)/L,\ \{H(a)-H(h)\}/L,\ \log(1+c_i)\bigr],
\end{equation}
where $H$ is Shannon entropy, $L=\log C$ for $C>1$ clauses, and $L=1$ for a single clause. We standardize inputs, apply ridge penalty 1 to the weights, and leave the intercept unpenalized. The final intercept is 0.07477 and the four weights are $(0.05334,0.00902,-0.04770,-0.01945)$. Training uses nonempty RxR development contacts, first in leave-one-building-out folds and finally in all development buildings. The frozen threshold is $1.293\times10^{-6}$ predicted success per extra byte. On SAGE the matched global probability is 0.7920, and the four cost-bin probabilities are 1, 1, 0.4616 and 0.

Held-out-building development predictions favored the selector over globally matched random by $0.74\pp$ (pointwise interval $[+0.05,+1.41]$). Matching spending within bins reduced that difference to $0.27\pp$ $[-0.16,+0.71]$. The thumbnail completed 172 development tasks and the selector 169. Model fitting and threshold selection end in the development buildings. The external estimates in \fig{f:perturb} then use the frozen policies.

On the held-out SAGE scenes the selector sent all 231 and 150 available images in the two lowest size bins, 42 of 86 in the third bin and none of 25 in the highest. The globally matched random policy transmitted 389.65 images in expectation. The selector and the full-image policy both completed 244 routes. Selection reduced total application bytes by 20.8\% relative to transmitting every full image. The fixed thumbnail saved a further 24.3\% relative to the selector. Sending every full image added 13,190,784 bytes over P across the 492 contacts.

{\RaggedRight \textit{Reproducible estimates and computational cost.}
The release includes 3,920 autonomous departures in \texttt{anc/paired.csv}. The extended \texttt{anc/paired\_v5.csv} adds each solo outcome, navigation metrics and run provenance. Each row identifies the cohort, scene, episode, contact, start distances, packet outcomes and measured bytes. SAGE rows also give the frozen send decision. A blank distance encodes an unavailable geodesic. The maps in \fig{f:cases} use each receiver's initial heading and a shared physical scale. The shaded 3\,m region consists of navigable cells whose goal distance, queried on the same navigation mesh at 0.06\,m spacing, is at most 3\,m. Its shape can follow an obstacle even when a projected straight line looks short.\par}

Four RTX~4090 GPUs supplied 64.07 GPU-worker hours for the formal designed-scene evaluation (64.90 including integration and reserved bootstrap time). Feature extraction and instruction alignment averaged 0.071\,s and 2.327\,s per contact. Average receiver-rollout times were 13.32, 12.72 and 12.35\,s under \arm{P}, \arm{P+I} and \arm{P+T}, respectively.

The registered E1--E4 evaluation produced 16,408 receiver continuations: 10,078 under new conditions and 6,330 same-batch baseline reruns. GPU allocations for these experiments, including E1 frame preparation and the frozen E4 progress estimator, totaled 108.593 GPU hours. A separate 160-continuation reproducibility check used 0.962 GPU hours.
\section{Reproducibility and Registration}\label{app:registration}
\textit{Fresh policy decisions.}
From the sorted autonomous contacts in each cohort we sampled 50 SAGE and 30 RxR handoffs, using seed zero separately for each cohort. Each episode was restored from its recorded receiver pose, camera observation and budget. On SAGE we regenerated decisions under the progress packet and progress plus image. The two arms scored 22 and 30 successes in the original run and again in the repeat. On RxR, actions and actions plus image scored 12 and 14 successes in both runs. All 160 position paths, terminal distance states and success indicators matched exactly. Two RxR arm--episode pairs ended at a location disconnected from the goal in both runs. The archive contains the terminal positions but omits the final orientations. We therefore use new within-batch controls for the revision comparisons. The check consumed 0.962 GPU-worker hours across four RTX~4090 GPUs.

\textit{Registration and analysis units.}
The revision hypotheses and complete-pair analysis were committed locally on November 22, 2026, at 19:08:19 Pacific standard time, after the retrospective trajectory analysis and before any receiver used a new packet. The registration commit is

{\small\noindent\texttt{cd5b046cab2fe30df6ae56100eeb0555485fde11}.}

Each of the four comparisons has one primary estimand. For autonomous SAGE handoffs, N1 compares a correctly matched image with a mismatched one, and N2 adds the image to a constant first-clause index. N3 compares the two reference-cue packets over all 1,122 controlled RxR placements, retaining the unknown code when the reference cannot identify a unique clause. N4 adds the image to the estimated progress cue at 1,136 controlled SAGE placements outside the goal radius. We report pooled paired differences with 98.75\% two-sided intervals for this four-comparison family. Additional autonomous estimates include every departure. Controlled placements estimate receiver performance under placement. Whole scenes are resampled in SAGE, and routes within the same building are resampled in RxR. Reporting includes the realized denominators, adverse flips and the fraction of RxR references that uniquely identify a clause.

Table~\ref{t:registered_pairs} gives the outcome of both packets on each eligible route. All four primary comparisons have complete pairs.

\textit{Reference progress.}
The frozen alignment identified a unique clause for 48 of the 1,122 controlled placements (4.28\%). The other 1,074 carry index zero. N3 therefore measures the image effect at controlled handoffs when the reference progress index is mostly unknown. For the 48 known indices we report only descriptive rates and paired outcome counts. We filed this interpretation at 17:10:43 Los Angeles time on November 23 (\texttt{d16f9a6}). At 17:12:33 all four output directories had zero completed reference-plus-image records. The first completed episode in that arm began its policy calls at 17:19:01.944 and ended at 17:19:07.466. Its complete record was observed at 17:19:07.589.

Among the 48 uniquely labeled placements, the reference cue alone yielded 20 successes and the cue plus image yielded 22. Six succeeded only with the image, four only without it, 16 under both packets and 22 under neither. Among the 1,074 placements with an unknown index, successes rose from 335 to 359. The secondary paired difference in that group was $+2.23\pp$ (pointwise 95\% CI $[-0.84,+5.35]$).

{\RaggedRight The two coverage rates use different annotations. FGR2R~\cite{fgr2r} supplies sub-instruction viewpoint ranges for the R2R training routes. A unique published subpath-to-clause match labeled 171/179 prefixes. RxR supplies word times for a human recording and discrete panorama visits~\cite{rxr}. Our independently executed continuous prefix also needs a unique spatial anchor with unambiguous word-to-clause support. For 801 routes the endpoint lacked a unique match on its ordered reference edge.\par}

The frozen RxR segmentations contain one clause for 824 routes, multiple clauses for 292, and no clause for six. An exact lookup across 16 existing segmentation files recovered none of those six. One admissible index in a single-clause instruction supplies no timing evidence. In one two-clause example (instruction 121415), the lead ended 0.088\,m from the matched panorama. During that visit the guide spoke words from both clauses. The recording places the transition at 22.2\,s, but the robot's separately executed prefix supplies no corresponding recording time. The released fields and frozen segmentations therefore do not identify a reference index at every controlled endpoint.

\textit{Paired changes when the progress index is unknown (post hoc).}
At the 1,074 controlled RxR placements with an unknown index, A succeeded on 339 routes and $P_{\rm ref}$ on 335 (34 successes only with A and 30 only with $P_{\rm ref}$). With the image the counts were 371 and 359 (44 and 32 only, respectively). At the other 48 placements the counts were 16 and 20 without the image, and 24 and 22 with it. Across all 1,122 no-image pairs, 319 succeeded under both packets, 36 under each packet alone, and 731 under neither. In the 222 autonomous SAGE handoffs with an unknown estimated index, A and P scored 85 and 81 (14 and 10 only) and A+I and P+I scored 95 and 103 (6 and 14 only). We reconstructed first-call token sequences from archived packets and run settings with the available evaluator and tokenizer. The runs did not log live token arrays or visual tensors. On all 1,068 nonempty unknown RxR prefixes and all 222 unknown SAGE handoffs, the semantic prompt appended \emph{Current sub-instruction is unknown.} (seven extra tokens). The ordered visual slots were the same, but their token positions shifted. The two image packets contained byte-identical JPEGs on all 1,116 nonempty controlled RxR prefixes and all 492 autonomous SAGE handoffs.

\textit{Implementation deviation at empty prefixes.} Six controlled RxR placements had no lead actions. Their native action packets were empty and the first decision used one current view. A four-byte reference packet encoded zero actions and an unknown index, invoking an empty action turn with the current view in two slots. Neither image condition carried a JPEG. On the six placements, both action-packet arms scored six successes and both reference-packet arms scored five. All six contributed zero to the registered N3 paired difference. The v4 implementation uses the native single-view branch for zero actions, an unknown index and no image. This correction concerns dispatch on future replays. The reported counts are the original 1,122 measured pairs.

\textit{Secondary N2 intervals.}
After observing E3a, we changed the interval reported for team success per departure from 95\% to 98.75\%. We give both levels for paired terminal-error and SPL differences among realized contacts. The four registered primary estimands retain their original definitions.

\textit{Progress and placement secondary measurements.}
Of 179 R2R calibration prefixes, 171 had unique FGR2R labels. In N2 the constant index plus image rescued 100 routes and reversed 56 successes. The mean image increment was 26.81\,kB. Scene-equal team success rose from 7.20\% to 8.80\%. At contact, mean terminal error fell from 4.22 to 3.65\,m (paired 95\% CI for the difference $[-0.85,-0.28]$\,m and 98.75\% CI $[-0.94,-0.21]$\,m). SPL rose from 0.254 to 0.335 (paired 95\% CI $[+0.045,+0.116]$ and 98.75\% CI $[+0.035,+0.126]$).

At the 1,136 controlled outside-radius SAGE placements in 94 scenes, N4 recorded 367 progress-only and 384 progress-plus-image successes. The secondary scene-equal effect was $+1.85\pp$ (pointwise 95\% CI $[-1.23,+4.90]$). Mean terminal errors were 5.51 and 5.41\,m, with a paired difference of $-0.09$\,m (95\% CI $[-0.27,+0.08]$).

{\RaggedRight\textit{Post-registration supplementary Solo comparison (S1).}
The local registration was committed at 08:42:01 on November 24, 2026 (Los Angeles, PST), before extracting previously unscored Solo outcomes or generating a continuation for S1. Earlier lead logs and team outcomes already existed. The first archived Solo terminal score was generated at 08:47:39 PST. The commit is \texttt{a4e1a28ed98c4d51500b668c9237028e668558e3}. S1 is separate from both prespecified families. It compares each packet with the Solo reference in two separate populations. The contact estimate pairs only realized handoffs within each cohort. The departure estimate pairs all routes within that cohort. Both use two-sided pointwise 95\% intervals from 10,000 resamples (whole scenes in SAGE and paired routes within each RxR building). The recovery fraction uses paired ratios of the same bootstrap draws only where the interval for Solo minus actions lies wholly above zero.\par}

Before formal continuation, all 20 sampled prefixes (12 SAGE and eight RxR) reproduced every one of 970 archived positions at the original float32 precision. The frozen lead proposed the archived executed actions throughout these prefixes. At contact, 14 of the 20 had actions left in the current decoded block. Solo retained them and continued from the lead's actual state. The replay gate passed at 08:55:36 PST. Of the 3,095 complete noncontact logs, 3,008 ended in STOP and 87 exhausted the action budget. All 825 original contact logs ended at first contact. We continued every one of them, preserving unconsumed actions from the current policy block on 611 routes. The first new contact continuation ended at 08:56:03 PST. Both cohorts met the recovery-ratio rule, with zero undefined denominators in 10,000 draws per cohort.

\textit{Concordance across the formal baseline reruns.}
All 6,330 rerun--archive comparisons agreed on success, terminal position, the full recorded position sequence, terminal geodesic value or unreachable state, step count, stopping, packet length and episode seed. All 6,270 pairs with finite terminal distances matched numerically, and the remaining 60 were unreachable in both runs. The 3,444 comparable SAGE action sequences also agreed exactly. The RxR archive omits action sequences, and the archives omit terminal orientations for these comparisons. These comparisons involve 4,362 distinct archived cohort--protocol--route--condition records. Of those, 984 SAGE records were each rerun in three batches and 3,378 were rerun once.
\section{Application Delivery at Handoff}\label{app:links}
\begin{table*}[t]\caption{Application delivery for representative recorded packet sizes. Each cell summarizes 200 transfers. Times are median / 95th percentile, in ms.}\label{t:netem}
\centering\small\setlength{\tabcolsep}{3pt}\begin{tabular*}{\textwidth}{@{\extracolsep{\fill}}lrrrrrr@{}}\toprule
Packet & Bytes & TCP 6 Mbit/s & TCP 250 kbit/s & UDP 250 kbit/s & UDP loss (\%) & Workstation--server ACK \\ \midrule
Count & 2 & 10.2 / 10.2 & 27.2 / 27.3 & 26.5 / 26.7 & 1.0 & 56.2 / 81.1 \\
A & 6 & 10.2 / 10.2 & 27.4 / 27.5 & 26.6 / 26.8 & 2.5 & 56.8 / 82.5 \\
P & 8 & 10.1 / 10.2 & 27.4 / 27.6 & 26.7 / 26.9 & 2.0 & 53.0 / 76.6 \\
R1 & 1,093 & 11.6 / 11.7 & 62.2 / 62.3 & 61.4 / 61.6 & 0.5 & 68.9 / 97.3 \\
R2 & 2,079 & 13.0 / 21.1 & 95.9 / 122.8 & 94.1 / 94.2 & 6.0 & 71.0 / 106.6 \\
R3 & 3,681 & 15.2 / 22.0 & 149.2 / 233.3 & 146.4 / 146.6 & 3.5 & 76.7 / 103.3 \\
R4 & 7,169 & 20.1 / 30.0 & 265.1 / 296.8 & 260.2 / 260.4 & 9.0 & 78.7 / 105.2 \\
R5 & 11,794 & 26.6 / 44.6 & 421.5 / 504.5 & 411.5 / 411.6 & 15.5 & 83.6 / 122.1 \\
P+T & 16,081 & 32.6 / 56.9 & 565.0 / 600.1 & 551.9 / 552.1 & 21.0 & 88.7 / 127.6 \\
P+I & 26,777 & 49.4 / 76.7 & 922.1 / 1014.5 & 902.9 / 903.1 & 35.5 & 109.8 / 159.3 \\
\bottomrule\end{tabular*}
\par\smallskip\parbox{0.96\textwidth}{\footnotesize TCP and UDP columns measure simulated one-way application delivery in private Linux network namespaces; UDP quantiles condition on receipt. The last column reports a physical two-host measurement over an existing SSH/TCP connection, timed at the sender until the receiver acknowledged the complete payload.}
\end{table*}

We selected the observed payload size nearest each message type's median. Ties were resolved by episode identifier. Each size was transmitted 200 times per link. For the emulation, sender and receiver ran in separate Linux network namespaces connected by a 1,500-byte-MTU virtual Ethernet pair. Both directions were shaped. The fast link used 6~Mbit/s, 20~ms round-trip delay and 1\% packet loss per direction. The low-rate link used 250~kbit/s, 50~ms round-trip delay and 2\% loss. TCP retained its connection across transfers. In the UDP comparison one application datagram was sent per trial, with IP fragmentation and a 3~s delivery deadline. We timed first send to receipt of the final payload byte against a common monotonic clock.

For a physical two-host measurement, the same payload lengths were interleaved in a fixed random order over the existing workstation--server connection. The workstation timed its send through receipt of an application acknowledgment. This round-trip endpoint includes the return path. Table~\ref{t:netem} reports it separately from one-way simulated delivery.
\section{Image Encoding and Paired Outcomes}\label{app:e2}
\begin{table*}[t]\caption{Encoded bytes, paired outcomes and receiver trajectories for seven versions of the same lead image at 492 SAGE handoffs.}\label{t:e2_all}
\centering\fontsize{8}{9.5}\selectfont\setlength{\tabcolsep}{2pt}
\begin{tabular*}{\textwidth}{@{\extracolsep{\fill}}llrrcrrrr@{}}\toprule
\multicolumn{9}{@{}l}{\textbf{A. Application bytes and outcomes at contact.} Each row uses the same 492 handoffs.} \\[2pt]
Packet & \shortstack[l]{JPEG pixels,\\quality} & \shortstack[r]{Total bytes\\mean / median} & \shortstack[r]{Additional bytes\\mean / median} & \shortstack{Successes\\(SR, \%)} & \shortstack[r]{Gained /\\lost} & \shortstack[r]{$\Delta$SR (pp)\\{[95\% CI]}} & \shortstack[r]{kB/net\\gain} & \shortstack[r]{Expected UDP\\$\Delta$SR (pp)} \\ \midrule
P & --- & 8 / 8 & 0 / 0 & 198 (40.24) & --- & ref. & --- & ref. \\
R1 & 64$\times$48, 50 & 1,088 / 1,093 & 1,080 / 1,085 & 198 (40.24) & 72 / 72 & $0.00\;[-4.58,+4.65]$ & --- & $0.00$ \\
R2 & 128$\times$96, 50 & 2,083 / 2,080 & 2,075 / 2,072 & 217 (44.11) & 83 / 64 & $+3.86\;[-1.58,+9.20]$ & 53.7 & $+3.63$ \\
R3 & 160$\times$120, 75 & 3,714 / 3,685.5 & 3,706 / 3,677.5 & 220 (44.72) & 82 / 60 & $+4.47\;[-0.59,+9.51]$ & 82.9 & $+4.32$ \\
R4 & 320$\times$240, 50 & 7,240 / 7,168.5 & 7,232 / 7,160.5 & 235 (47.76) & 90 / 53 & $+7.52\;[+2.47,+12.58]$ & 96.2 & $+6.84$ \\
R5 & 384$\times$384, 50 & 11,945 / 11,824 & 11,937 / 11,816 & 225 (45.73) & 85 / 58 & $+5.49\;[+0.65,+10.23]$ & 217.5 & $+4.64$ \\
P+T & 320$\times$240, 90 & 16,080 / 16,067.5 & 16,072 / 16,059.5 & 240 (48.78) & 101 / 59 & $+8.54\;[+2.80,+14.09]$ & 188.3 & $+6.74$ \\
P+I & 384$\times$384, 90 & 26,819 / 26,790 & 26,811 / 26,782 & 244 (49.59) & 100 / 54 & $+9.35\;[+4.08,+14.62]$ & 286.8 & $+6.03$ \\
\bottomrule\end{tabular*}\vspace{5pt}
\begin{tabular*}{\textwidth}{@{\extracolsep{\fill}}lrrrrrr@{}}\toprule
\multicolumn{7}{@{}l}{\textbf{B. Team outcomes per departure and navigation after contact.}} \\[2pt]
Packet & \shortstack[r]{Team SR\\(\%)} & \shortstack[r]{$\Delta$ team SR (pp)\\{[95\% CI]}} & \shortstack[r]{$\Delta$ mean terminal error (m)\\{[95\% CI]}} & SPL & \shortstack[r]{Outside OSR\\(\%)} & \shortstack[r]{Median receiver\\actions} \\ \midrule
P & 7.13 & ref. & ref. & 0.252 & 64.79 & 38 \\
R1 & 7.16 & $+0.03\;[-0.77,+0.84]$ & $-0.01\;[-0.29,+0.26]$ & 0.250 & 61.97 & 36 \\
R2 & 7.70 & $+0.57\;[-0.40,+1.57]$ & $-0.28\;[-0.59,+0.04]$ & 0.285 & 66.20 & 34 \\
R3 & 7.86 & $+0.73\;[-0.15,+1.64]$ & $-0.29\;[-0.58,+0.01]$ & 0.296 & 65.49 & 32 \\
R4 & 8.42 & $+1.29\;[+0.41,+2.20]$ & $-0.40\;[-0.69,-0.11]$ & 0.318 & 66.90 & 32 \\
R5 & 8.03 & $+0.90\;[+0.05,+1.74]$ & $-0.35\;[-0.64,-0.05]$ & 0.309 & 66.20 & 32 \\
P+T & 8.52 & $+1.39\;[+0.35,+2.42]$ & $-0.46\;[-0.75,-0.17]$ & 0.321 & 64.79 & 32 \\
P+I & 8.77 & $+1.64\;[+0.72,+2.60]$ & $-0.50\;[-0.80,-0.19]$ & 0.325 & 66.20 & 32 \\
\bottomrule\end{tabular*}
\par\smallskip\parbox{0.98\textwidth}{\footnotesize Differences use the same-run progress packet (P). Each interval covers one comparison. Scenes are the resampling unit. Team SR is the scene-equal average over 95 scenes and 2,798 departures. Navigation error and SPL use all 492 contacts. Mean terminal error for P was 4.20\,m. Outside OSR measures entry into the goal region among the 142 receivers starting outside. Gained and lost count opposing success flips. Efficiency divides total additional bytes by net additional successes. R1 has a zero denominator. One kB is 1,000 bytes. Expected UDP $\Delta$SR multiplies the measured handoff gain by the delivery fraction at the representative size for that encoding (200 single-datagram transfers at 250\,kbit/s, 2\% per-fragment loss, Table~\ref{t:netem}). The projection assumes that delivery is independent of route-level gain within each encoding and that P is available on loss. Only point estimates are given for this projection.}
\end{table*}

All seven variants use the lead frame immediately before the shared action block. The actions, estimated progress clause and receiver view are fixed. R1--R4 and P+T resize the archived $640\times480$ RGB frame directly using bicubic interpolation. R5 and P+I start from the frozen encoder's square model view, denormalized and rounded to RGB before JPEG encoding. Each packet appends its encoded image to the same progress message. The receiver decodes the JPEG and applies its usual visual preprocessing. Table~\ref{t:e2_all} distinguishes total application bytes from the increment over the 8-byte progress packet.

For each encoding we divide the sum of extra bytes over paired contacts by the number of routes gained minus those lost. For the $320\times240$ quality-50 thumbnail this gives 3,557,931 bytes for 37 net additional successes, or 96.2\,kB each. The full model view uses 13,190,784 extra bytes for 46 net additional successes, or 286.8\,kB each. The seven encoding comparisons have pointwise 95\% intervals. They are separate from the four registered primary contrasts in Appendix~\ref{app:registration}.
\section{Image Substitution at Recorded Handoffs}\label{app:substitution}
Each substitution preserves the stored receiver view, instruction, remaining actions and packet fields preceding the image. The donor JPEG is copied intact from another contact. The donor was from the same scene on 486 of 492 SAGE handoffs and all 321 eligible RxR handoffs. In the other six SAGE pairs the registered size-bin rule allowed a different scene. Most comparisons therefore retain scene information while varying which route supplied the image. Figure~\ref{f:e2_rate} locates the measured SAGE donor cost alongside the seven encodings.

\begin{table}[!htbp]
\caption{Fidelity of the pre-block image. Reproduction compares the encoded re-render with the received JPEG after decoding. Encoding loss compares the original pre-JPEG model RGB with that same received image. Entries are cohort medians. PSNR is in dB.}\label{t:frame_fidelity}
\centering\small
\begin{tabular*}{\columnwidth}{@{\extracolsep{\fill}}lrrrrr@{}}\toprule
 & & \multicolumn{2}{c}{Reproduction} & \multicolumn{2}{c}{Encoding loss}\\
\cmidrule(lr){3-4}\cmidrule(lr){5-6}
Cohort & Pairs & PSNR & SSIM & PSNR & SSIM\\
\midrule
SAGE & 492 & $\infty$ & 1.000 & 41.12 & 0.984\\
RxR & 321 & $\infty$ & 1.000 & 38.69 & 0.968\\
\bottomrule\end{tabular*}
\end{table}

\begin{table*}[!t]\caption{Navigation at the same handoffs under image substitution and retiming. Byte counts include the complete application packet.}\label{t:substitution}
\centering\fontsize{8.1}{9.7}\selectfont\setlength{\tabcolsep}{3pt}
\begin{tabular*}{\textwidth}{@{\extracolsep{\fill}}llrrrrrrr@{}}\toprule
Cohort & Packet & \shortstack[r]{Successes /\\handoffs} & SR (\%) & \shortstack[r]{Total bytes\\mean / median} & \shortstack[r]{Median terminal\\error (m)} & SPL & \shortstack[r]{Outside\\OSR (\%)} & \shortstack[r]{Median\\actions} \\ \midrule
SAGE & Progress only & 198 / 492 & 40.24 & 8 / 8 & 3.77 & 0.252 & 64.79 & 38 \\
SAGE & Matched pre-block & 244 / 492 & 49.59 & 26,819 / 26,790 & 3.04 & 0.325 & 66.20 & 32 \\
SAGE & Mismatched & 218 / 492 & 44.31 & 26,560 / 26,044.5 & 3.40 & 0.266 & 59.86 & 35.5 \\
SAGE & At contact & 210 / 492 & 42.68 & 25,708 / 25,303 & 3.37 & 0.265 & 58.45 & 35 \\
SAGE & Earlier & 238 / 492 & 48.37 & 26,366 / 25,826 & 3.09 & 0.323 & 60.56 & 30.5 \\
RxR & Actions only & 136 / 321 & 42.37 & 6 / 6 & 4.14 & 0.291 & 55.91 & 76 \\
RxR & Matched pre-block & 161 / 321 & 50.16 & 33,557 / 32,559 & 2.56 & 0.401 & 65.00 & 63 \\
RxR & Mismatched & 147 / 321 & 45.79 & 33,451 / 32,364 & 3.25 & 0.347 & 58.18 & 63 \\
RxR & At contact & 141 / 321 & 43.93 & 33,391 / 32,244 & 3.38 & 0.318 & 59.55 & 68 \\
\bottomrule\end{tabular*}
\par\smallskip\parbox{0.98\textwidth}{\footnotesize Every image is compared with its same-batch controls. SAGE packets retain executed actions and estimated instruction progress. Outside OSR measures entry among receivers starting outside the goal region (142 in SAGE). RxR image packets retain the actions. RxR rows cover 321 nonempty handoffs, with 310 finite terminal distances, 310 valid SPL values and 220 outside receivers.}
\end{table*}

\begin{table}[!htbp]
\caption{Paired success outcomes for the four registered comparisons. Gained counts routes successful only with the comparison image. Lost counts routes successful only with its reference packet. N1 replaces a donor image. N2--N4 add an image to a short message. Figure~\ref{f:perturb} gives the corresponding intervals.}\label{t:registered_pairs}
\centering\small
\begin{tabular*}{\columnwidth}{@{\extracolsep{\fill}}lrrrrr@{}}\toprule
Contrast & Routes & Both & Gained & Lost & Neither \\
\midrule
N1 & 492 & 159 & 85 & 59 & 189 \\
N2 & 492 & 145 & 100 & 56 & 191 \\
N3 & 1,122 & 220 & 161 & 135 & 606 \\
N4 & 1,136 & 245 & 139 & 122 & 630 \\
\bottomrule\end{tabular*}
\end{table}

\begin{figure*}[!t]
\centering\includegraphics[width=0.98\textwidth]{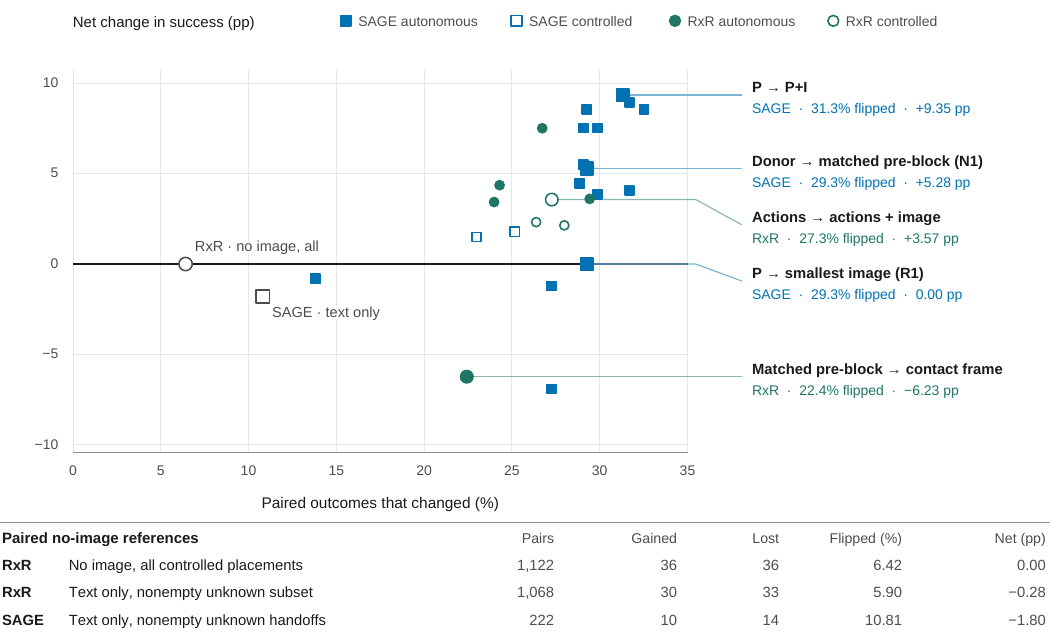}
\caption{Total changes in outcome and net gains for paired packet comparisons. For a pair of conditions, a gain is a route that succeeds only under the comparison packet, and a loss succeeds only under the reference. The horizontal coordinate counts both directions and the vertical coordinate subtracts losses from gains, each per 100 paired handoffs or placements. Thus a point can have many flips even at zero net gain. For image comparisons, squares denote SAGE and circles RxR. Filled symbols show autonomous handoffs and open symbols controlled placements. Two gray open markers add no-image references. The accompanying table isolates the 1,068 nonempty unknown RxR prefixes whose reconstructed first prompts differ only in the unknown-index text. The complete RxR group also contains 48 known indices and six empty prefixes with a different first-turn layout.}\label{f:gross_flips}
\end{figure*}

\textit{Timing and payload details.}
Using the pre-block frame instead of the contact frame increased SAGE successes from 210 to 244 (pointwise 95\% CI for the increment $[+1.93,+11.86]\pp$) and RxR successes from 141 to 161 ($[+1.00,+11.48]\pp$). In the SAGE earlier-frame comparison (E1c), 475 of 492 histories permitted the full eight-action shift. All selected a different frame index. The earlier and pre-block frames gave 238 and 244 successes, respectively, with 64 routes successful only under the earlier frame and 70 only under the pre-block frame. The scene-equal team difference was $-0.32\pp$ (95\% CI $[-1.23,+0.53]$). Mean additional matched and donor payloads were 26.81 and 26.55\,kB in SAGE, and 33.55 and 33.44\,kB in RxR.

\textit{Similarity to the receiver's current view (exploratory).}
We compared each saved transmitted JPEG with the receiver's actual current RGB frame at the same handoff. We used the encoder's bicubic $384\times384$ view and mean three-channel Gaussian SSIM~\cite{ssim_wang} ($11\times11$ window, $\sigma=1.5$, population covariance). Mean SSIM in SAGE was 0.486 at contact, 0.455 before the last action block and 0.449 eight actions earlier (492 matched routes). In RxR it was 0.421 at contact and 0.402 before the block (321 nonempty histories). The contact frame was closer than the pre-block frame on 334/492 SAGE and 206/321 RxR handoffs. These descriptive comparisons measure image overlap. They do not establish that overlap caused the navigation gains.

\textit{Render and encoding controls.}
We replayed the lead's archived actions from its original pose to recover the camera state immediately before the last shared action block. For a history of $n$ actions this is RGB frame $\max(0,n-4)$. The reference is the JPEG actually transmitted to the receiver. E1b renders the contact pose, whereas E1c starts from an earlier stored RGB frame. Reconstructing the pre-block frame lets us check rendering at a known target independently of changing capture time. We apply the frozen image preprocessing, integer rounding and quality-90 encoding to the re-render, then measure PSNR and structural similarity (SSIM)~\cite{ssim_wang} on decoded 8-bit RGB. SSIM averages the three channel scores using a Gaussian window ($\sigma=1.5$) and population covariance. Identical pixels have infinite PSNR and are counted separately from finite values. Re-encoding the archived RGB without rendering supplies a separate control for preprocessing and JPEG serialization.

We recovered SAGE camera orientations through the original sensor-quaternion conversion and rendered with the original Gaussian scene. RxR used its native Habitat renderer. All 813 replays reproduced the stored position trajectory and the original pre-block RGB pixels. After encoding and decoding, all 813 images also matched the received JPEG pixel for pixel (Table~\ref{t:frame_fidelity}). Directly re-encoding the archived RGB reproduced the transmitted bytes in every case.

In SAGE, the receiver succeeded only with the matched image on 85 routes and only with the donor on 59. Both arms succeeded on 159 and both failed on 189. Across all 2,798 departures, the secondary scene-equal gain was $0.93\pp$ (95\% CI $[-0.06,+1.93]$). Across the 492 paired handoffs, the matched image reduced mean terminal navigation error by 0.37\,m (matched minus donor $-0.37$\,m, 95\% CI $[-0.68,-0.04]$).

In RxR, the matched image alone succeeded on 46 routes and the donor alone on 32. Both succeeded on 115 and both failed on 128. These 321 nonempty histories form the comparison in \fig{f:perturb}. At 12 other contacts the empty histories give identical receiver inputs and outcomes across packets (eight successes and four failures). Including these gives 169/333 successes with the matched image and 155/333 with the donor, a gain of $4.20\pp$ (95\% CI $[-0.91,+9.38]$). Across all 1,122 departures, including 789 without contact, the same 14 net additional successes give $+1.25\pp$ $[-0.27,+2.76]$. The pre-block image gained 20 successes relative to the contact-time frame, or $+1.78\pp$ $[+0.27,+3.30]$ across departures. Among the 310 pairs with finite terminal distances in both arms, matched minus donor mean terminal error was $-0.25$\,m $[-0.99,+0.50]$. All RxR intervals here are pointwise 95\% intervals from resampling routes within building.

Across the same 321 nonempty RxR handoffs, the four counts in ascending order were 136 for the actions, 141 for actions with the contact-time image, 147 for actions with the donor image, and 161 for actions with the matched pre-block image (42.37, 43.93, 45.79 and 50.16\%). Including the 12 empty histories gives 144, 149, 155 and 169 successes out of 333. The order describes these counts. The paired comparisons in \fig{f:perturb} supply the intervals.

\textit{Post hoc combination across cohorts.}
Over the observed mixture of 492 SAGE and 321 RxR handoffs, the matched image yielded 405/813 successes (49.82\%) and the donor 365/813 (44.90\%). In 131 pairs only the matched image succeeded, in 91 only the donor succeeded, in 274 both succeeded and in 317 both failed. The pooled gain was $4.92\pp$ (exploratory 95\% CI $[0.90,8.95]$). We fixed the cohort weights at 492/813 and 321/813. Each of 10,000 replicates resampled paired SAGE scenes and paired RxR routes within buildings using independent random streams, then combined the two estimates with those weights. Each comparison retains its cohort's short packet and same-batch donor. This observed-mixture estimate complements the registered SAGE content comparison ($+5.28\pp$, 98.75\% CI $[-1.96,12.36]$).

\textit{Changed outcomes and net success.}
The smallest E2 image changed the result of 144 of 492 handoffs with zero net gain. When the packet included a constant first-clause index, adding the full image changed 156 of the same 492 handoffs, gaining 100 and losing 56. Figure~\ref{f:gross_flips} separates how often each packet changes a route's outcome from how often those changes favor it.
\fi
\section*{Acknowledgment}
The InteriorGS views in Figs.~\ref{f:handoff} and~\ref{f:cases}\ifextended{} and the Appendix~\ref{app:sender} frame sequences\fi{} use data provided by SpatialVerse Research Team, Manycore Tech Inc.~\cite{interiorgs}. OpenAI Codex assisted with \ifextended experiment execution, analysis code, \fi manuscript drafting and preparation of figures from recorded data. Claude (Anthropic) supplied the figure and table design brief and editorial feedback.
\bibliographystyle{IEEEtran}
{\RaggedRight\bibliography{references,revision/references_v3}}
\end{document}